\documentclass{bmvc2k}

\title{Evaluating Perceptual Distance Models by Fitting Binomial Distributions to Two-Alternative Forced Choice Data}

\addauthor{Alexander Hepburn}{alex.hepburn@bristol.ac.uk}{1}
\addauthor{Raul Santos-Rodriguez}{enrsr@bristol.ac.uk}{1}
\addauthor{Javier Portilla}{javier.portilla@csic.es}{2}
\addinstitution{
School of Engineering Mathematics \\
University of Bristol \\
Bristol, United Kingdom
}
\addinstitution{
Instituto de Optica \\
 Consejo Superior de Investigaciones Cient\'ificas (CSIC)\\
 Madrid, Spain
}

\runninghead{Hepburn et al.}{Evaluating Perceptual Distances with 2AFC Data}

\usepackage{amsmath}
\usepackage{amssymb}
\usepackage{mathtools}
\usepackage{amsthm}
\usepackage{subcaption}

\usepackage[capitalize,noabbrev]{cleveref}

\theoremstyle{plain}

\theoremstyle{definition}

\theoremstyle{remark}

\usepackage{microtype}
\usepackage{graphicx}
\usepackage{booktabs}
\usepackage{multirow}
\usepackage{booktabs}
\usepackage{hyperref}

\def \xv{{\bf x}}

\newcommand{\ud}{\mathrm{d}}

\begin{document}

\maketitle

\begin{abstract}
The Two Alternative Forced Choice (2AFC) paradigm offers advantages over the Mean Opinion Score (MOS) paradigm in psychophysics (PF), such as simplicity and robustness. However, when evaluating perceptual distance models, MOS enables direct correlation between model predictions and PF data. In contrast, 2AFC only allows pairwise comparisons to be converted into a quality ranking similar to MOS when comparisons include shared images. In large datasets, like BAPPS, where image patches and distortions are combined randomly, deriving rankings from 2AFC PF data becomes infeasible, as distorted images included in each comparisons are independent. To address this, instead of relying on MOS correlation, researchers have trained ad-hoc neural networks to reproduce 2AFC PF data based on pairs of model distances -— a black-box approach with conceptual and operational limitations. This paper introduces a more robust distance-model evaluation method using a pure probabilistic approach, applying maximum likelihood estimation to a binomial decision model. Our method demonstrates superior simplicity, interpretability, flexibility, and computational efficiency, as shown through evaluations of various visual distance models on two 2AFC PF datasets.
\end{abstract}

%-------------------------------------------------------------------------
\section{Introduction}
\label{sec:intro}

% The fundamental problem of replicating the ability of the human visual system to compare images is increasingly important in computer vision. 
% Many \emph{perceptual distance models} have attempted to capture the behaviour of the visual system, whether through extracting and comparing structures present in images~\citep{Wang04, wang2003multiscale}, modelling mechanisms that are present in the visual pathway~\citep{laparra2016, Hepburn2020perceptnet}, or more recently using neural networks pre-trained for classification to extract image features~\citep{zhang2018unreasonable, ding2020image}.

Replicating the human visual system’s ability to compare images is a fundamental challenge in computer vision. Many perceptual distance models aim to approximate this behaviour, whether by comparing image structures~\citep{Wang04, wang2003multiscale}, modelling visual mechanisms~\citep{laparra2016, Hepburn2020perceptnet}, or extracting features from classification-trained neural networks~\citep{zhang2018unreasonable, ding2020image}.

These models are typically evaluated using data from psychophysical experiments, often in a two-alternative forced choice (2AFC) format~\citep{fechner1948elements}. Participants are shown triplets composed by a reference and two distorted images, and asked which distorted image is closer to the original. Responses are sometimes aggregated into mean opinion scores (MOS) to rank distortions~\citep{tid2008-data, tid2013-data, live-data, csiq-data}, which requires careful control over which images are shown and shared across triplets. Alternatively, one can assume a probabilistic model of internal perceptual scores~\citep{thurstone1994law}.

Traditional datasets are collected under tightly controlled viewing conditions, yielding reliable but limited data. Recently, practitioners have dropped those constraints in favour of more reference images and a larger number of distortions using custom software tools to simplify the process \cite{CLIFFORD2025102225}. The Berkeley Adobe Perceptual Patch Similarity (BAPPS)~\citep{zhang2018unreasonable} contains judgements of 187,700 image patches using 425 distortions, far more than traditional datasets such as TID2013~\citep{tid2013-data}. BAPPS contains the raw perceptual judgements, as opposed to MOS, as it is infeasible to compute a ranking using triplets that do not share images~\citep{tsukida2011analyze}.
However, evaluating perceptual distance models using such data is non-trivial: a mapping from the distances between the two distorted images and the reference to the proportion of people who judged the first distorted image closer to the reference needs to be learnt. This mapping often ignores the number of human participant judgements per image triplet (which can vary), a piece of information that can be used to infer uncertainty on the judgement.  
%The CLIC dataset~\citep{CLIC2021} also releases raw perceptual judgements, although with a varying number of judgements per triplet.
%Datasets containing only the MOS for each given reference-distorted image pair are also missing some key information, such as the number of human participant judgements per image triplet which can be used to infer uncertainty over the judgement. 

Importantly, our goal is not to estimate perceived image distances per se, but to evaluate how well perceptual distance models explain 2AFC data. This requires mapping the pair of computed distances $(d_0, d_1)$ to the probability that the first distortion is judged closer. Existing approaches often rely on simplified assumptions about internal scores~\citep{thurstone1994law}, and typically ignore uncertainty from varying numbers of human ratings.

We follow relevant literature and model the stochastic nature of the perceptual judgement process using a binomial distribution~\citep{thurstone1994law, perez2017practical}. We compute distances using various perceptual models, and estimate the judgements' likelihood using a kernel-smoothed density estimation. This approach is computationally efficient, accounts for varying numbers of ratings, and avoids overconfidence in data-sparse regions. Evaluation metrics such as the likelihood of a judgement according to the binomial model are simple to compute, and can explicitly account for a different number of human judgements used in the experiment, even for individual triplets. We also compare with a small neural network trained to maximise the likelihood of the responses according to the binomial model. Our density-based model achieves similar performance with significantly fewer parameters and training time.

\section{Related Work}
\label{sec:related_work}
\paragraph{Perceptual Experiments} are performed using a number of methodologies, that differ in both the way the stimuli are presented, and the interaction the participant has with the experiment. We focus on the 2AFC experimental setup, used by popular datasets such as TID 2008 \& 2013~\citep{tid2008-data, tid2013-data} and BAPPS~\citep{zhang2018unreasonable}. However, TID and other datasets use 2AFC experiments to \emph{rank} distorted images, where the MOS is then calculated and only this is released. In contrast, BAPPS presents \emph{randomly composed} triplets to observers and releases the raw 2AFC judgements. The CLIC dataset~\citep{CLIC2021} also releases the 2AFC judgements, but in that case triplets have a variable number of judgements. An overview of the available visual perception datasets can be found in Appendix~\ref{ap:dataset}.

For MOS-based datasets, one can simply calculate correlations between the distance between each reference-distorted pair according to the model, and the MOS. However, with datasets like BAPPS, this is not possible. A common approach is to force the decision to be binary: \emph{according to the distance model, is the first distorted image closer to the reference than the second?} This ignores the number and distribution of judgements performed per triplet, wrongly equating a unanimous decision to one that is close to a tie. %, are the same.
\cite{zhang2018unreasonable} addressed this by mapping the distance pair $(d_0, d_1)$ to the proportion of observers preferring the first distortion, using a small neural network trained with cross-entropy loss. This also enables end-to-end optimisation of the perceptual model. However, this does not explicitly model the stochastic nature of the decision process, and evaluation is merely based on prediction accuracy. Comparing models fairly thus requires fitting a separate network for each distance function.

%An alternative methodology, for which the proposed method can be applied, is alternate forced choice experiments with a quadruple judgement, where participants are asked: \emph{is the difference between one pair greater than (or less than) the difference between the other pair?} This is refereed to as the method of quadruples~\citep{Kingdom10}. There are many more methods involving forced choice when presented with different stimuli~\citep{creelman1979auditory, bi2019four}, which also generalise to more modalities than just vision. However many of these methods are used to determine aspects such as sensitivity in a particular direction, and require different assumptions of the underlying psychometric function than that we present in this paper.

Our proposed method assumes we have access to 2AFC experimental results, although it generalises to any alternative forced choice experiments such as the method of quadruples~\citep{Kingdom10}. We focus on BAPPS, as in that database there is no relationship between triplets (no possible ranking), which is what the proposed method is designed for. We also report results on the CLIC dataset, where there is a variable number of judgements per triplet.

\paragraph{Perceptual Distances} in the traditional literature are hand-crafted based on vision science insights. Metrics like SSIM~\citep{Wang04} and MS-SSIM~\citep{wang2003multiscale} assess structural similarity—how humans perceive image structure. Others focus on \emph{error visibility}, comparing images in perceptual spaces using Euclidean distance. For example, the normalized Laplacian pyramid distance (NLPD)~\citep{laparra2016} learns transformations that reduce spatial redundancy. Neural networks have recently been used to extract features for perceptual judgments. LPIPS~\citep{zhang2018unreasonable} leverages features from classification networks, learning a weighted average aligned with human perception via cross-entropy minimization on 2AFC data. While effective, it lacks a psychophysical basis and cannot estimate judgment likelihoods. DISTS~\citep{ding2020image} extends this by incorporating texture similarity. PIM~\citep{bhardwaj2020unsupervised} learns representations that maximize mutual information between adjacent video frames, which are then used for distance computation. For the candidate distances that we wish to evaluate using the proposed method, we select a variety of traditional and deep learning-based metrics: Euclidean distance, NLPD~\citep{laparra2016}, SSIM~\citep{Wang04}, PIM~\citep{bhardwaj2020unsupervised}, LPIPS~\citep{zhang2018unreasonable} and DISTS~\citep{ding2020image}. 

\paragraph{Bayesian approaches to vision} such as Maximum Likelihood Difference Scaling (MLDS) ~\citep{maloney2003maximum}, fit perceptual difference scales by modelling human judgements as arising from Gaussian-distributed internal responses. The scales are optimised to maximise the likelihood of observer orderings and apply to continuous physical attributes like luminance or contrast. Thurstone~\citep{thurstone1994law} similarly assumed perceptual quality follows a Gaussian distribution, proposing various models based on psychometric assumptions. These models estimate Gaussian parameters by fitting observed judgements to a binomial distribution~\citep{tsukida2011analyze, silverstein2001efficient, jogan2014new}, with the mean reflecting internal scale differences.

In contrast to these classical approaches, we do not assume any internal response model here. Instead, we directly model the 2AFC decision as a binomial process, where the sole parameter is the probability of choosing one option over another. This framework allows us to estimate judgement probabilities for any number of responses. We can reliably estimate the fitness of every candidate perceptual distance model in a conceptually and computationally simple way, without adhering to model assumptions about subjective responses.

\section{Method}
\label{sec:method}
Let us consider psychophysical data from 2AFC experiments between a reference image and two degraded images, which are presented to the subjects~\citep{tid2008-data, tid2013-data, zhang2018unreasonable, CLIC2021}.
Each one of the $T$ triplets of images $(\xv_{ref}(t),\xv_0(t),\xv_1(t))$ receives a fixed number $M$ of responses. We define $n(t) \in [0, M]$ as the number of times $\xv_1(t)$ is deemed closer to $\xv_{ref}(t)$ than $\xv_0(t)$ according to the observers. We model $n(t)$ as a binomial random variable $n \sim \mathcal{B}(M, P)$, where $P$ is the probability of choosing $\xv_1$.

The underlying hypothesis is that there is a perceptual distance model $d$ that maps two images to their corresponding visual distance, such that the choice count $n$ is conditionally independent of $\{\xv_{ref},\xv_0,\xv_1\}$ given $(d(\xv_{ref},\xv_0),d(\xv_{ref},\xv_1))$. We also define the distance between the reference and first distorted as $d_0(t)=d(\xv_{ref}(t), \xv_0(t))$ and distance between the reference and second distorted as $d_1(t)=d(\xv_{ref}(t), \xv_1(t))$. The previous hypothesis is equivalent to stating that there is a function $f$, such that the binomial parameter $P$ controlling the probability of the choice is $P(\xv_{ref},\xv_0,\xv_1) = f(d_0,d_1)$.

We will refer to the function $f$ as $P(d_0, d_1)$ for simplicity, and the parameter corresponding to a particular triplet $t$ as $P(d_0(t), d_1(t))$. Given a set of psychophysical data from a 2AFC experiment, here we address the problem of measuring \emph{how well that candidate function fits the empirical data under the assumed binomial process}. At this point, it is important to note that we should not choose $P(d_0(t),d_1(t))$ simply as its most likely value ($n(t)/M$) according to that particular triplet $t$. Due to the noise in the decision process, we must consider other triplet results in the neighbourhood to obtain a smooth and consistent $P(d_0,d_1)$ function on the whole $(d_0,d_1)$ distance plane. Reliable estimation requires densely sampled data over this space.

\subsection{Density estimation}
We estimate the empirical densities $\{p(j,d_0,d_1), j = 1\dots M\}$, which model how likely are the events $\{(d_0,d_1),n=j\}$. As explained before, to obtain a continuous probability density function (PDF) from a set of discrete events, we use a Gaussian kernel integrating the corresponding samples on the distance plane. This is done separately for each $j$, ensuring that $\int p(j,d_0,d_1), \ud d_0\, \ud d_1= P(j) = 1/T\sum \delta(n(t)-j)$.
Then, we compute the overall density on the distance place as $p(d_0,d_1) = \sum_{j=1}^M p(j,d_0,d_1)$.
(Note that we can take advantage of the symmetry $p(d_0,d_1) = p(d_1,d_0)$, by enforcing it.)
Now we can estimate the conditional probabilities
\begin{equation}\label{eq:conditional}
    Pr(n=j|d_0,d_1) = \frac{p(j,d_0,d_1)}{p(d_0,d_1)},j=1\dots M.
\end{equation}
To ensure coverage of the whole $(d_0, d_1)$ plane in $[0, 1]$, we first marginally uniformise the training data (details in Appendix~\ref{ap:uniform}). We then construct a uniform grid over this space, for which we will estimate the conditional probabilities. Gaussian kernels of a suitable size, centered at each of the discrete points in the training set and evaluated on the grid, are then summed. This is applied separately to each of the sets $\{(d_0,d_1),n=j\}$, thus obtaining estimations of $p(j,d_0,d_1)$ for all $j$s. Finally, we compute their sum ($p(d_0,d_1)$) and apply Eq.~\ref{eq:conditional} to obtain the estimated conditional probabilities. According to the assumed model, $n(t)\sim \mathcal{B}(M,P(d_0(t),d_1(t)))$, we know the theoretical binomial probability of each $n$ as a function of $P(d_0,d_1)$ and $M$, that we term $Pr_{\mathcal{B}}(n=j;M,P(d_0,d_1))$. From the above, we estimate $\hat P(d_0,d_1)$, the probability parameter maximising the log-likelihood of the observations for every $(d_0,d_1)$: $\hat P(d_0,d_1) = \arg\max_{P} L(d_0,d_1;M,P)$. The distribution $\hat{P}(d_0, d_1)$ that maximises the likelihood of the empirical data originating from the assumed binomial distribution is given by
\begin{equation}\label{eq:maximised_likelihood}
    \hat P(d_0,d_1) =\frac{1}{M} \sum_{j=1}^{M} {j \cdot Pr(n=j|d_0,d_1)},
\end{equation}
with a derivation in Appendix~\ref{ap:max_like}. It is important to note that the modelling of the decision-making process does not depend on $M$, but $M$ contributes to the empirical data gathered. This allows us to evaluate with a variable number of judgements with the same underlying learned distribution.

\begin{figure}[h]
    \centering
    \subfigure[]{%
        \includegraphics[width=0.08\textwidth]{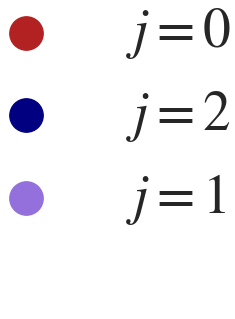}
    }
    \subfigure[]{%
        \includegraphics[width=0.2\textwidth]{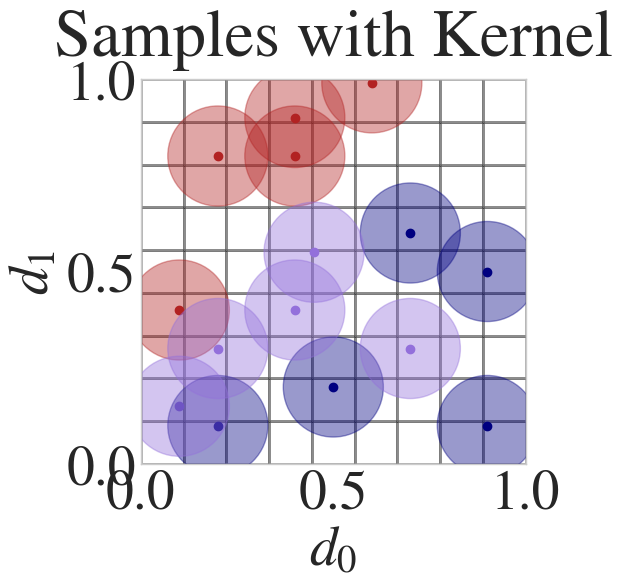}
    }
    \subfigure[]{%
        \includegraphics[width=0.15\textwidth]{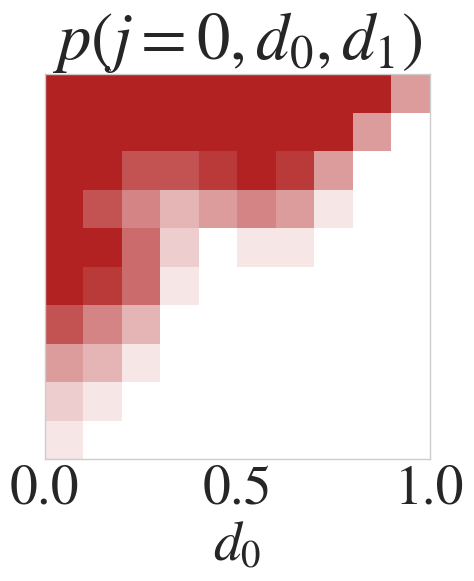}
    }
    \subfigure[]{%
        \includegraphics[width=0.15\textwidth]{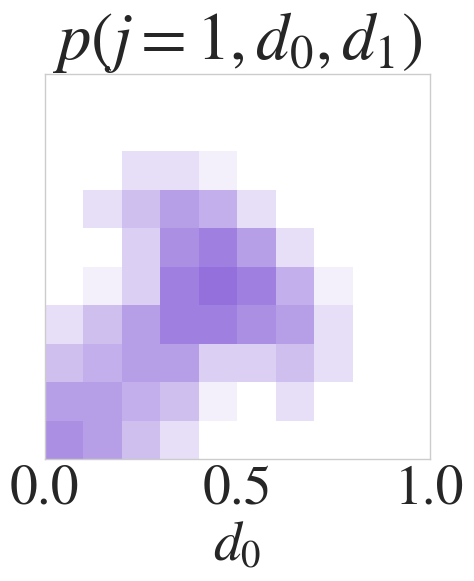}
    }
    \subfigure[]{%
        \includegraphics[width=0.15\textwidth]{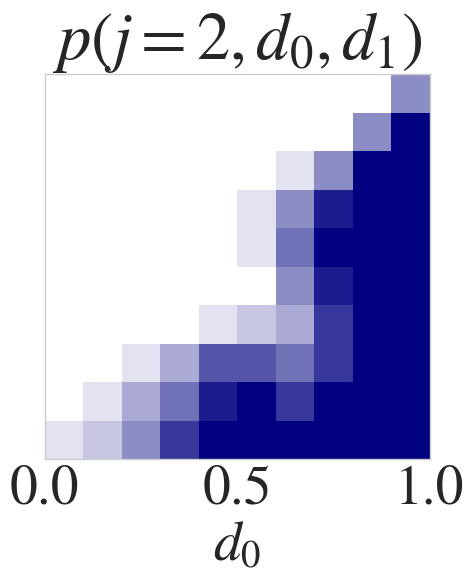}
    }
    \subfigure[]{%
        \includegraphics[width=0.15\textwidth]{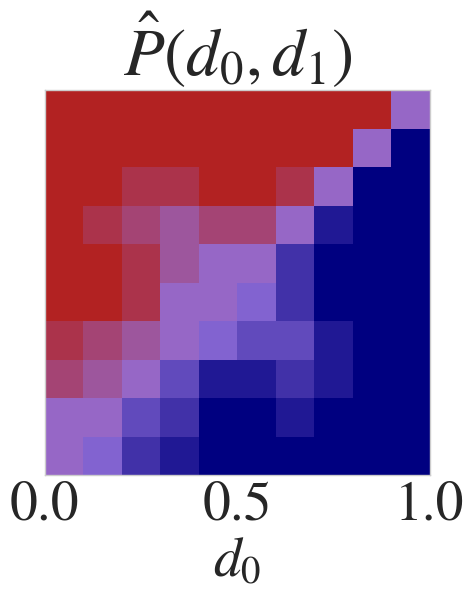}
    }
    \caption{An example using 5 data points, $M=2$ for each $j=\{0, 1, 2\}$. (a) Samples with a Gaussian kernel applied (the circle represents the standard deviation) and a $10\times10$ grid for which we estimate %the conditions in 
    Eq.~\ref{eq:conditional}. (b), (c) and (d) are the estimated conditional distributions for each value of $j$ with $M=2$, and (e) is the distribution after maximum likelihood estimation according to Eq.~\ref{eq:maximised_likelihood}.}
    \label{fig:example_fit}
\end{figure}

Finally, we average the local (in $(d_0,d_1)$) optimal log-likelihood \\ $\hat L(d_0,d_1) = L(d_0,d_1;M,\hat P(d_0,d_1))$ on the whole $(d_0,d_1)$ plane providing a global evaluation of how well the binomial distribution with the $d$ distance fits the empirical 2AFC data. Fig.~\ref{fig:example_fit} shows an example of the method, using a small amount of data points.

%Note that we can generalise this result to account for a different number of judgements $M$ per triplet $t$, $M_t$, by transforming the $M_t$ judgements of each given triplet into binary judgements ($M=1$) on $M_t$ identical triplets. Now we can construct the conditional distributions $p(j=\{0, 1\}, d_0, d_1)$, where the $P$ that maximises the likelihood of empirical probabilities is simply $\hat{P}(d_0, d_1) = p(j=1, d_0, d_1)/p(d_0, d_1)$. For example, where $M=2$ and both participants select $\mathbf{x_0}$ ($n(t)=2)$, this is equivalent to two individual judgements preferring $\mathbf{x_0}$ in two identical triplets. 

\subsection{Maximum-likelihood using a neural network}
\citet{zhang2018unreasonable} train a neural network to predict $n(t)/M$ via cross-entropy minimisation. For comparison, we use a neural network to estimate $\hat{P}(d_0,d_1)$ by maximising the likelihood of the training judgements, assuming a binomial distribution allowing for a more complex function heading to the previously stated modelling assumptions.

Given training samples $\{d_0(t), d_1(t), n(t)\}$ for $t=1,...,T$, we minimise the negative log-likelihood according to our binomial distribution;
\begin{equation}\label{eq:nll_likelihood}
    \begin{aligned}
        \text{NLL}&(\{n(t), t=1\dots T\}, \hat{P}, M) = 
        -\frac{1}{T}\sum_{t=0}^T \log{Pr_{\mathcal{B}}(n(t);M,\hat{P}(d_0(t), d_1(t)))}.
    \end{aligned}
\end{equation}
To enforce symmetry, we extend the training set with mirrored examples: $\{d_0(t), d_1(t), n(t)\}$ and $\{d_1(t), d_0(t), M-n(t)\}$, resulting in $2T$ samples.

\subsection{Evaluation Metrics}\label{sec:evaluation_metrics}
We propose two types of metrics to evaluate how well each distance model fits the binomial: (1) decision agreement with the model, and (2) log-likelihood of judgments under the learned binomial distribution.

If we apply the criterion of comparing the most likely outcome of the binomial distribution $\mathcal{B}(M,P)$, which is $\lfloor (M+1)P \rfloor$~\citep{feller1991introduction}, with the actual judgement $n(t)$, both normalised to $M$ for each triplet, we obtain a percentage agreement between our probability model $\mathcal{B}(M,\hat P(d_0(t),d_1(t)))$ and the human judgements $n(t)$, 
yielding the following expression:
\begin{equation}\label{eq:agreement}
    \begin{aligned}
        \text{AJ}(\{n(t), t=1\dots T\}, \hat{P}, M) =
        100 - \frac{100}{T}\sum_{t=1}^{T}{\left|\frac{\lfloor (M+1)\hat{P}(d_0(t), d_1(t))\rfloor-n(t)}{M}\right|}.
    \end{aligned}
\end{equation}
Note that the $M$ used here corresponds to the judgement $n(t)$, using the already fit probability model $\hat{P}$. The $M$ used in the evaluation can be different from the one used in the estimation of $\hat{P}$, for example, when a test set contains a different number of judgements.

To maximise the agreement for the learned model, one can generate random samples $\hat{n}(t)$ from our learned distribution $\hat{n}(t) \sim \mathcal{B}(M, \hat{P}(d_0(t), d_1(t)))$, and see the agreement between the empirical and the simulated judgements, according to our learned model. This provides a reference for the case that the observed judgements follow the fitted binomial model exactly, as reported in Appendix~\ref{ap:eval}.

Due to our assumptions of an underlying probability model for the decision process, we can also evaluate log-likelihoods of judgements according to the estimated binomial parameter. Rather than just measure the percentage of judgements that agree with our model, we can evaluate the negative log-likelihood of the empirical data according to the learned binomial model given by Eq.~\ref{eq:nll_likelihood}.

% \begin{equation}\label{eq:nll_binomial}
%     \text{NLL}(\{n(t), t=1\dots T\}, \hat{P}, M) = -\frac{1}{T} \sum^T_{t=1} \log{Pr_{\mathcal{B}}(n(t);M,\hat{P}(d_0(t), d_1(t)))}    
% \end{equation}

Finally, a regularly used metric is the agreement purely between the decisions; \emph{does the perceptual distance select the same distorted image as the humans?} For a set of experiments, this is given by
\begin{equation}\label{eq:2afc}
    \begin{aligned}
        \text{2AFC} = \frac{1}{T}\sum_{t=1}^T &\big[\hat{p}(d_0(t), d_1(t))\cdot \frac{n(t)}{M} + 
        (1 - \frac{n(t)}{M})\cdot(1 - \hat{p}(d_0(t), d_1(t)))\big],
    \end{aligned}
\end{equation}
where $n(t)$ is the number of humans that selected the first distorted image as closer to the reference, and $\hat{p}\in \{0, 1\}$ is the preference of the perceptual distance model, indicating if $d_0 > d_1$ (1) or otherwise (0).

\subsection{Extension to different number of judgements}
We extend the previous method to variable number of judgements per triplet $M_t$, by transforming the judgements into binary judgements ($M=1$) on $M_t$ identical triplets. Now we can construct the conditional distributions $p(j=\{0, 1\}, d_0, d_1)$, where the maximum likelihood estimate is $\hat{P}(d_0, d_1) = p(j=1, d_0, d_1)/p(d_0, d_1)$. For example, if $M=2$ and both responses favour $\mathbf{x_0}$ ($n(t)=2)$, this is equivalent to two individual judgements. The evaluation metrics remain consistent, but using $M_t$ for each judgement rather than $M$.

\section{Experiments}
\label{sec:experiments}
We can apply our likelihood model to six candidate perceptual distances, using BAPPS training set, containing triplets $\{\mathbf{x}_{ref}, \mathbf{x}_0, \mathbf{x}_1\}$ which are patches of size $64\times64$ and $M=2$ judgements (ie. $n(t)/M \in \{0.0, 0.5, 1.0\}$). Evaluation is performed on the validation set with $M=5$. Since we are directly estimating $P(d_0,d_1)$, which does not depend of the number of judgements used in the experiment $M$, we can train using $M=2$ and evaluate using $M=5$. After uniformisation the range of $\{d_0, d_1\}$ is $[0, 1]$, and we smooth discrete $\{d_0,d_1\}$ samples using a Gaussian kernel with a constant width. The width $\sigma$ can be set by the user -- in our experiments, we found that the method was robust to a wide range of $\sigma$ values that depended on the amount of data that covers the plane, i.e., less data requires a larger $\sigma$. For the BAPPS dataset, we set $\sigma=\frac{1}{44}$. We use a $20\times20$ grid to estimate the conditionals in Eq.~\ref{eq:conditional}. Sec.~\ref{sec:robustness} examines the proposed method's robustness to changes in these hyperparameters ($\sigma$ and grid density). Source code can be found at \url{https://github.com/alexhepburn/2afc_binomial}.

For comparison, we train a neural network with the same structure as the one from \citet{zhang2018unreasonable} to estimate $\hat{P}(d_0, d_1)$. The architecture is two fully connected layers, with 32 channels, followed by a 1-channel fully connected layer with sigmoid activation. We train for 5 epochs, batch size 128 and Adam (lr= $0.001$). Note that for each epoch, each data point is seen twice as we try to implement (but not strictly enforce) symmetry in $\hat{P}$.

\section{Results}
\label{sec:results}

\begin{figure}[h]
    \centering
    \subfigure[Density Estimation]{%
        \includegraphics[width=0.99\textwidth]{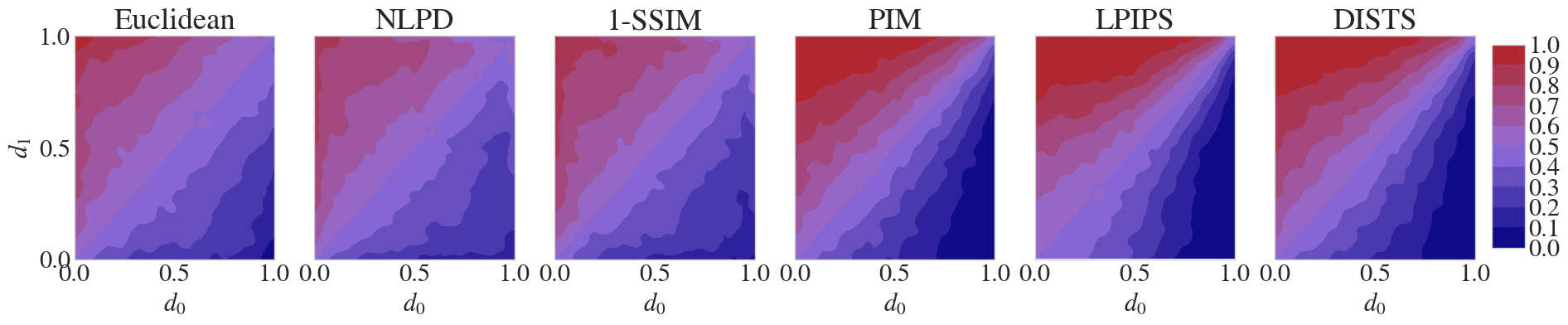}
    }
    \subfigure[Neural Network]{%
        \includegraphics[width=0.99\textwidth]{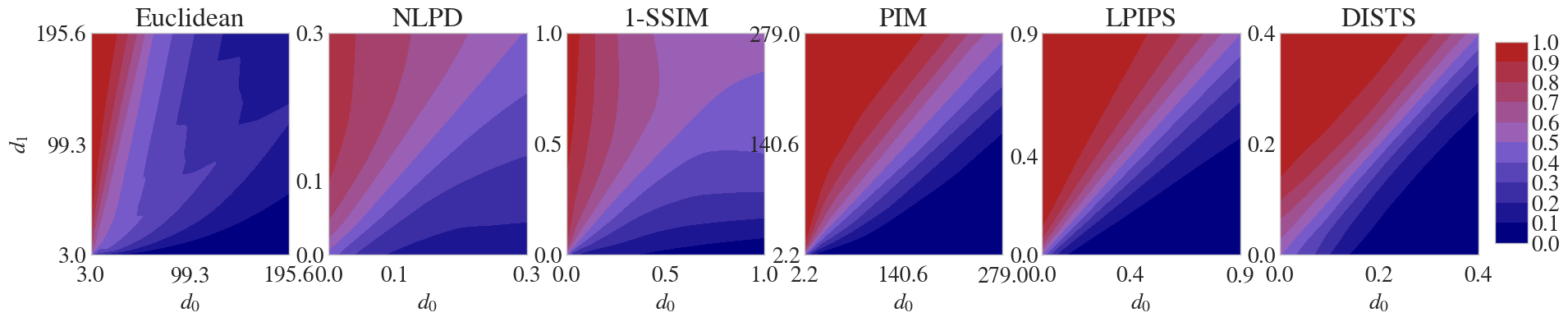}
    }
    \vspace{0.1cm}
    \caption{Binomial parameter $P$ estimated from the BAPPS training set for different perceptual distance models using (a) Density estimation and (b) Neural network.}
    \label{fig:decision_surface}
\end{figure}

\begin{table}[htb]
\centering
% \footnotesize
\resizebox{\textwidth}{!}{%
\begin{tabular}{llllllll}
\hline
 & \multicolumn{1}{c}{Measure} & \multicolumn{1}{c}{Euclidean} & \multicolumn{1}{c}{NLPD} & \multicolumn{1}{c}{SSIM} & \multicolumn{1}{c}{PIM} & \multicolumn{1}{c}{LPIPS} & \multicolumn{1}{c}{DISTS} \\ \hline
\multirow{3}{*}{\begin{tabular}[c]{@{}l@{}}Density\\ Estimation\end{tabular}} & $\text{AJ}(n, \hat{P}, M)$ \%$\uparrow$ & $75.89\pm0.00$ & $75.82\pm0.00$ & $76.17\pm0.00$ & $82.12\pm0.00$ & $82.43\pm0.00$ & $81.34\pm0.00$ \\
 & NLL$(n, \hat{P}, M)$ $\downarrow$ & $1.86\pm0.00$ & $1.86\pm0.00$ & $1.84\pm0.00$ & $1.49\pm0.00$ & $1.46\pm0.00$ & $1.53\pm0.00$ \\
 & 2AFC Score \% $\uparrow$ & $63.46\pm0.00$ & $63.31\pm0.00$ & $63.65\pm0.00$ & $70.09\pm0.00$ & $68.94\pm0.00$ & $69.12\pm0.00$ \\ \hline
\multirow{3}{*}{\begin{tabular}[c]{@{}l@{}}Neural\\ Network\end{tabular}} & $\text{AJ}(n, \hat{P}, M)$ \%$\uparrow$ & $75.76\pm0.09$ & $75.79\pm0.08$ & $76.18\pm0.18$ & $82.07\pm0.13$ & $82.55\pm0.03$ & $81.40\pm0.03$ \\
 & NLL$(n, \hat{P}, M)$ $\downarrow$ & $1.87\pm0.00$ & $1.86\pm0.00$ & $1.84\pm0.01$ & $1.50\pm0.01$ & $1.46\pm0.00$ & $1.53\pm0.00$ \\
 & 2AFC Score \% $\uparrow$ & $62.43\pm1.31$ & $63.78\pm1.20$ & $64.19\pm1.37$ & $70.31\pm1.34$ & $70.02\pm1.04$ & $68.81\pm0.49$ \\ \hline
\end{tabular}%
}
\vspace{0.3cm}
\caption{Evaluation metrics on the BAPPS validation set ($M=5$). Reported is the mean and standard deviation over 10 runs. Lower NLL is better. Note that for PIM, we had to disregard one of the runs as the training became unstable.}
\label{tab:results_bapps}
\end{table}

We evaluate the metrics in Sec.~\ref{sec:evaluation_metrics} for the validation set of BAPPS. Table~\ref{tab:results_bapps} contains the results after optimising separate binomial models for each of our six distance models. The first three models (Euclidean, NLPD and SSIM) achieve similar results in agreement measures and negative log-likelihoods. The three deep-learning-based models (PIM, LPIPS and DISTS) achieve superior performance. %Additionally, the negative log-likelihoods for the simulated judgements $\text{NLL}(\hat{n}(t), \hat{P}, M)$, the minimum possible, is lower (better), as expected in Appendix~\.
Evaluation metrics on simulated judgements from the learned models and the training set can be seen in Appendix~\ref{ap:bapps_results}. Table~\ref{tab:distance_only} compares the 2AFC score using just the raw distance, vs using $\hat{P}(d_0, d_1)$ (as in Table 1). Most distances show just a slight increase in score, apart from Euclidean and LPIPS. Note that LPIPS parameters have been optimised to explicitly minimise the cross-entropy between the network outputs and the ground truth proportion $n(t)/M$.

As our density estimation method is deterministic, the standard deviation of the evaluation metrics in Table~\ref{tab:results_bapps} is $0.0$. However, with neural networks, there is a slight deviation between runs due to the stochastic nature of training, with a standard deviation of the 2AFC score above $1.0$ for all metrics except DISTS. This also caused us to exclude one of the runs using PIM, as training became unstable. In addition, we can visualise $\hat{P}(d_0, d_1)$ for different distances, to see the separation or amount of uncertainty along the diagonal, where $d_0$ and $d_1$ are similar. Fig.~\ref{fig:decision_surface} shows $\hat{P}(d_0, d_1)$, revealing that the three deep-learning-based metrics offer more certainty in the top left ($d_0<<d_1$) and bottom right ($d_0>>d_1$). As the distances $\{d_0, d_1\}$ increase (top right), the uncertainty region grows smaller, reflecting how humans perceive sharper differences in images far from the reference. The traditional distances display a wider area around $0.3<\hat{P}(d_0, d_1)<0.7$, indicating higher uncertainty on the decision. In comparison, the neural networks display non-symmetric unpredictable behaviour in Euclidean, NLPD and SSIM, as symmetry is not strictly enforced, and overconfidence in judgements when using PIM, LPIPS and DISTS. Additionally, the uncertainty increases when the distances are larger for all distance models except DISTS; the opposite of what is observed in humans.

\begin{table}[htb]
\centering
\footnotesize
\resizebox{\textwidth}{!}{%
\begin{tabular}{@{}lllllll@{}}
\toprule
2AFC Score \% $\uparrow$        & Euclidean & NLPD  & SSIM  & PIM   & LPIPS* & DISTS \\ \midrule
Distance Only     & $63.38\pm0.00$     & $62.71\pm0.00$ & $63.11\pm0.00$ & $69.26\pm0.00$ & $69.40\pm0.00$ & $68.48\pm0.00$ \\
Density Estimation & $63.46\pm0.00$ & $63.31\pm0.00$ & $63.65\pm0.00$ & $70.09\pm0.00$ & $68.94\pm0.00$ & $69.12\pm0.00$ \\ 
Neural network & $62.43\pm1.31$ & $63.78\pm1.20$ & $64.19\pm1.37$ & $70.31\pm1.34$ & $70.02\pm1.04$ & $68.81\pm0.49$ \\ \bottomrule
\end{tabular}}
\vspace{0.3cm}
\caption{2AFC Scores (Eq.~\ref{eq:2afc}) on the BAPPS test set, where distance only is checking if $d_0 > d_1$, and $\hat P(d_0,d_1)$ is our approach. *LPIPS has been optimised to replicate decisions in the BAPPS training set using a neural network mapping $(d_0, d_1)$ to $n(t)/M$.}
\label{tab:distance_only}
\end{table}

We compare training time and number of parameters between our density estimation method and using neural networks. The parameters in density estimation are the grid for which we estimate $\hat{P}(d_0, d_1)$ for, we use a $20\times20$ grid resulting in 400 parameters versus 1281 in the neural network. It is also significantly faster to train; across 10 training runs, the minimum training time (giving a quickest time possible) for density estimation was 4.5s and for neural networks 43.3s, despite neural networks utilising a GPU. More details can be found in Appendix~\ref{ap:performance}.

\begin{figure}[htb]
        \centering
        \includegraphics[width=.5\textwidth]{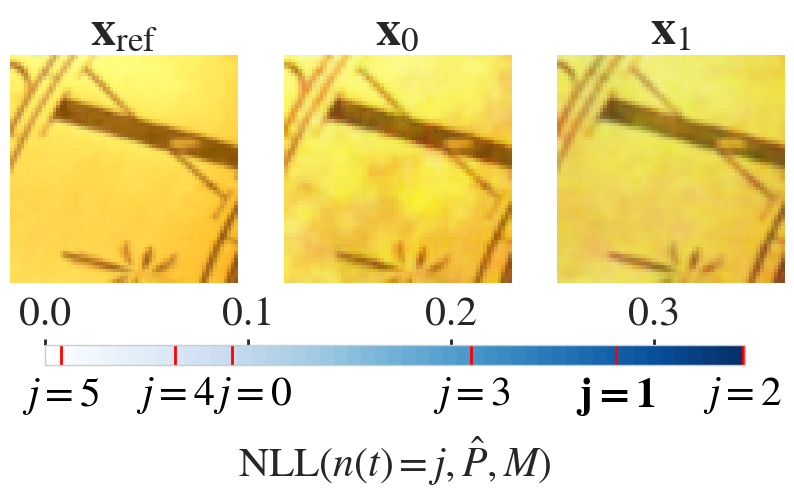}
        \vspace{0.2cm}
        \caption{Example of valuating the negative log-likelihood $j=[0, 5]$ according to DISTS for a triplet from the BAPPS test set where one image $\mathbf{x}_0$ is close to the reference $\mathbf{x}_{\text{ref}}$. $j=0$ denotes 5 participants selecting $\mathbf{x}_0$ as closer to $\mathbf{x}_{\text{ref}}$. For NLL, White is more likely and blue is less likely.}
        \label{fig:int_eg1}
\end{figure}

\paragraph{Interpretability}
One advantage of explicitly modelling the decision is that we can evaluate the negative log-likelihood of different $n(t)$ values using Eq.~\ref{eq:nll_likelihood}, allowing users to query the model for a given perceptual distance model. This holds for both the proposed density estimation and the neural network when optimised for log-likelihood. Fig~\ref{fig:int_eg1} shows the negative log-likelihood of a different number of observers $j=[0, 5]$ preferring the far right image over the middle, according to the density estimation method. More examples can be found in Appendix \ref{ap:int-examples}. Practitioners can see where a certain perceptual distance model fails, assess the probability of the decision, and how the likelihood will change with changes in $(d_0, d_1)$ and the number of judgements $M$.

% \begin{figure}[htb]
%     \centering
%     \includegraphics[width=0.49\textwidth]{figs/interpretable.png}
%     \caption{An example of querying the density estimation probability model for different number of people that prefer $\mathbf{x}_0$ over $\mathbf{x_1}$, $j=[0, 5]$, values for a triplet from the BAPPS test set, according to DISTS where the true $j$ is shown in bold. Shown above the colour bar is negative log-likelihood, where white is more likely and blue is less likely.}
%     \label{fig:nll_triplet}
% \end{figure}

% \begin{figure}[h]
%     \centering
%     \includegraphics[width=0.4\textwidth]{figs/interpretable_new.png}
%     \vspace{0.3cm}
%     \caption{An example of evaluating the density estimation probability model for different $j=[0, 5]$, for a triplet from the BAPPS test set, according to DISTS where the true $j$ is shown in bold. Shown above the colour bar is negative log-likelihood, where white is more likely and blue is less.}
%     \label{fig:nll_triplet}
% \end{figure}

\begin{figure}[h]
    \centering
    \includegraphics[width=.7\textwidth]{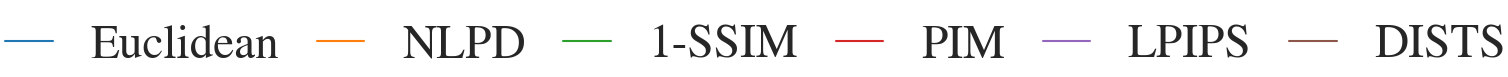} \\
    \subfigure[Width of the Gaussian kernel $\sigma$]{%
        \includegraphics[width=0.7\textwidth]{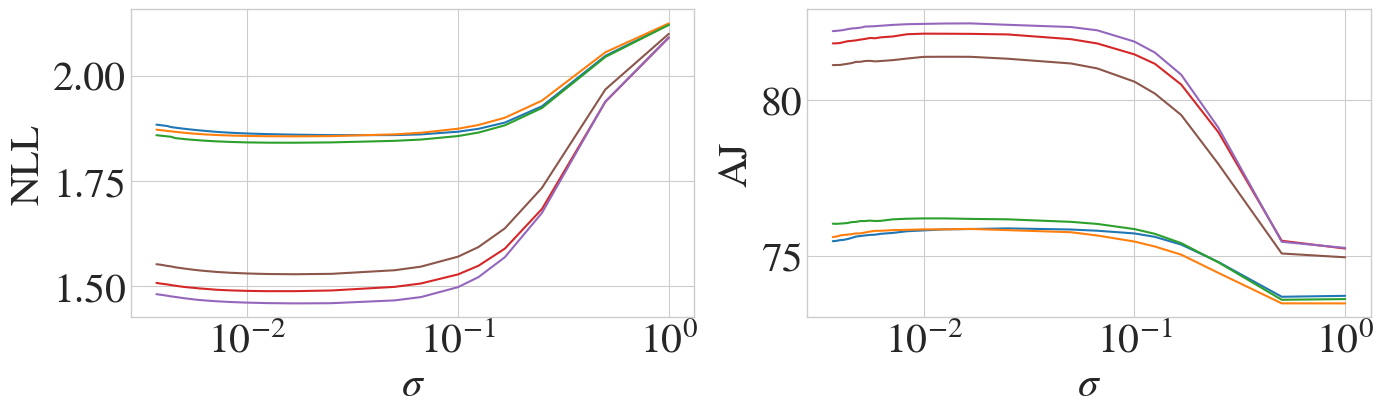}
        \label{fig:w_robustness}
    }
    \subfigure[Resolution of the grid used to estimate $\hat{P}(d_0, d_1)$.]{%
        \includegraphics[width=0.7\textwidth]{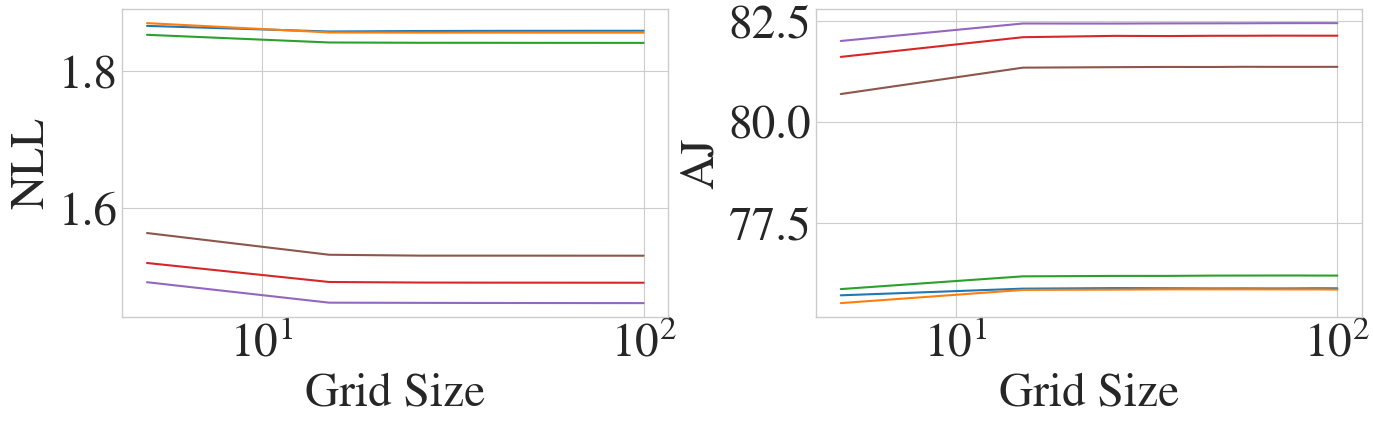}
        \label{fig:nll_robustness}
    }
    \vspace{0.15cm}
    \caption{Variability of the density estimation method with relation to (a) the width of the Gaussian kernel $\sigma$ and (b) the number of partitions in the grid used to estimate $\hat{P}(d_0, d_1)$. In each subplot, left is NLL (Eq.~\ref{eq:maximised_likelihood}), right is AJ (Eq.~\ref{eq:agreement}).}
    \label{fig:robustness}
    % \vspace{-3mm}
\end{figure}

\paragraph{Robustness}\label{sec:robustness}
Our method has two main hyperparameters; Gaussian kernel width $\sigma$, and the grid resolution for estimating $\hat{P}(d_0, d_1)$. We optimise $\hat{P}$ on the training set ($M=2$) by varying each hyperparameter and evaluating NLL and AJ on the test set ($M=5$). To isolate the effect of $\sigma$, we fix the grid to $100 \times 100$ and vary $\sigma \in [0.0037, 1]$. Fig.~\ref{fig:w_robustness} shows that a large $\sigma$ (heavy smoothing) results in performance loss. Fig~\ref{fig:decision_surface}) allows the user to decide a $\sigma$ that determines the smoothness of the estimated distribution $\hat{P}(d_0, d_1)$. Unlike neural networks, this method offers transparent, predictable control over model behaviour via interpretable hyperparameters.

The grid size determines resolution of $\hat{P}(d_0, d_1)$. Fig.~\ref{fig:example_fit} (a) shows an example of a $10\times10$ grid on data in $[0, 1]$. We test grid sizes in $[5, 100]$ (Fig.\ref{fig:nll_robustness}) showing that a too coarse grid does not allow proper estimation, and a plateau beyond $20$. Smaller grid size reduces computation but may limit accuracy, depending on the spread of $\{d_0, d_1\}$ after marginal uniformisation. For data similarly distributed to BAPPS, performance remains stable across a wide range of grid sizes, and this behaviour is consistent across different distance metrics.

\subsection{Different number of judgements}
We also present results with the  CLIC 2021 dataset~\cite{CLIC2021}, which includes triplets with varying numbers of judgements $M_t$. Judgements using an ``anchor'' (i.e. one of the distorted images is the reference image) are removed as they distort the uniformisation transform due to edge effects ($d_0 \, \text{or} \, d_1 = 0.0$). We use the oracle set (119,901 triplets with $M_t=\{1, 2\}$) for training and the validation set (4807 triplets with $M_t \in[1, 10]$) for evaluation. The distribution of $M_t$ can be seen in Appendix~\ref{ap:clic}. 

On CLIC, Euclidean, NLPD, and SSIM struggle to distinguish between distorted images, as seen in Table~\ref{tab:results_clic}, due to the non-uniformity in the CLIC measurements, despite the marginal uniformisation, seen in Appendix~\ref{ap:clic}, where these three distances display a large amount of uncertainty in the surface. More data could alleviate this issue. PIM, LPIPS, and DISTS display much more expected behaviour, with PIM achieving a 2AFC score of $0.7305$. Neural networks show similar trends, with higher variance for Euclidean and NLPD.

\section{Conclusions}
We introduce a method to evaluate perceptual distance models on 2AFC data, assuming a simple observer decision process. Using kernel smoothing and marginal uniformisation, we estimate PDFs over $(d_0, d_1)$ and apply maximum likelihood to infer choice weights. Applied to BAPPS, our method matches prior rankings while offering richer metrics. It is robust to hyperparameter changes, requires no additional training, and achieves neural network-level performance with fewer parameters, faster training, and interpretable hyperparameters. 

Unlike traditional ranking-based datasets with costly curated triplets, our method assumes random triplet selection and requires many samples for reliable PDF estimation—aligning with modern crowd-sourced datasets. It can be extended to optimize distance models via log-likelihood, as in LPIPS. A key consideration is the mismatch between the number of judgments in training and testing data: while the binomial parameter itself is independent of the number of judgements used, the data used to estimate it is not.

\begin{table}[t!]
\centering
\footnotesize
\resizebox{\textwidth}{!}{%
\begin{tabular}{llllllll}
\hline
 & \multicolumn{1}{c}{Measure} & \multicolumn{1}{c}{Euclidean} & \multicolumn{1}{c}{NLPD} & \multicolumn{1}{c}{SSIM} & \multicolumn{1}{c}{PIM} & \multicolumn{1}{c}{LPIPS} & \multicolumn{1}{c}{DISTS} \\ \hline
\multirow{3}{*}{\begin{tabular}[c]{@{}l@{}}Density\\ Estimation\end{tabular}} & $\text{AJ}(n, \hat{P}, M)$ \%$\uparrow$ & $44.27\pm0.00$ & $44.70\pm0.00$ & $44.85\pm0.00$ & $74.02\pm0.00$ & $74.03\pm0.00$ & $75.99\pm0.00$ \\
 %& $\text{AJ}(\hat{n}, \hat{P}, M)$ \%$\uparrow$ & 23.73 & 24.38 & 24.01 & 23.00 & 23.67 & 23.12 \\
 & NLL$(n, \hat{P}, M)$ $\downarrow$ & $0.72\pm0.00$ & $0.72\pm0.00$ & $0.73\pm0.00$ & $0.62\pm0.00$ & $0.61\pm0.00$ & $0.58\pm0.00$ \\
 %& NLL$(\hat{n}, \hat{P}, M)$ $\downarrow$ & 1.47 & 1.48 & 1.45 & 1.31 & 1.29 & 1.32 \\
 & 2AFC Score \% $\uparrow$ & $42.94\pm0.00$ & $43.08\pm0.00$ & $43.62\pm0.00$ & $73.18\pm0.00$ & $73.14\pm0.00$ & $75.39\pm0.00$ \\ \hline
\multirow{3}{*}{\begin{tabular}[c]{@{}l@{}}Neural\\ Network\end{tabular}} & $\text{AJ}(n, \hat{P}, M)$ \%$\uparrow$ & $46.25\pm1.04$ & $45.86\pm3.01$ & $45.13\pm0.18$ & $73.89\pm0.08$ & $73.97\pm0.06$ & $75.87\pm0.05$ \\
 %& $\text{AJ}(\hat{n}, \hat{P}, M)$ \%$\uparrow$ & 23.74 & 24.84 & 10.97 & 26.17 & 23.00 & 23.06 \\
 & NLL$(n, \hat{P}, M)$ $\downarrow$ & $0.73\pm0.00$ & $0.72\pm0.00$ & $0.76\pm0.00$ & $0.62\pm0.01$ & $0.61\pm0.00$ & $0.58\pm0.00$ \\
 %& NLL$(\hat{n}, \hat{P}, M)$ $\downarrow$ & 1.48 & 1.48 & 1.52 & 1.33 & 1.28 & 1.31 \\
 & 2AFC Score \% $\uparrow$ & $45.01\pm1.09$ & $44.45\pm3.05$ & $43.85\pm0.18$ & $73.20\pm0.09$ & $73.14\pm0.06$ & $75.26\pm0.04$ \\ \hline
\end{tabular}}
\vspace{0.3cm}
\caption{Results on the CLIC test set ($M_t=[1, 10]$). Reported is the mean and standard deviation over 10 runs. Lower NLL is better.}
\label{tab:results_clic}
\end{table}

\pagebreak

\section{Acknowledgements}
This work was funded by UKRI Turing AI Fellowship EP/V024817/1, TAILOR, a project funded by the EU Horizon
2020 research and innovation programme under grant agreement no. 952215, by the Spanish Government grants PID2020-118071GB-I00 and PID2020 113596GB-I00, the VIS4NN project- Programa Fundamentos de la Fundación BBVA 2022, and by the TEC-2024/COM-322 (IDEAL-CV-CM) project from the Comunidad de Madrid.

\bibliography{ref}

\begin{thebibliography}{24}
\providecommand{\natexlab}[1]{#1}
\providecommand{\url}[1]{\texttt{#1}}
\expandafter\ifx\csname urlstyle\endcsname\relax
  \providecommand{\doi}[1]{doi: #1}\else
  \providecommand{\doi}{doi: \begingroup \urlstyle{rm}\Url}\fi

\bibitem[Bhardwaj et~al.(2020)Bhardwaj, Fischer, Ball{\'e}, and Chinen]{bhardwaj2020unsupervised}
Sangnie Bhardwaj, Ian Fischer, Johannes Ball{\'e}, and Troy Chinen.
\newblock An unsupervised information-theoretic perceptual quality metric.
\newblock \emph{Advances in Neural Information Processing Systems}, 33:\penalty0 13--24, 2020.

\bibitem[Clifford et~al.(2025)Clifford, Hepburn, Kleinlein, Vila-Tomás, Hernández-Cámara, Dauden, Lepora, Laparra, and Santos-Rodríguez]{CLIFFORD2025102225}
M.~Clifford, A.~Hepburn, R.~Kleinlein, J.~Vila-Tomás, P.~Hernández-Cámara, P.~Dauden, N.~Lepora, V.~Laparra, and R.~Santos-Rodríguez.
\newblock Iqm-vis: A user-centric python toolbox for visualising and evaluating image quality metrics.
\newblock \emph{SoftwareX}, 31:\penalty0 102225, 2025.
\newblock ISSN 2352-7110.
\newblock \doi{https://doi.org/10.1016/j.softx.2025.102225}.
\newblock URL \url{https://www.sciencedirect.com/science/article/pii/S235271102500192X}.

\bibitem[Ding et~al.(2020)Ding, Ma, Wang, and Simoncelli]{ding2020image}
K.~Ding, K.~Ma, S.~Wang, and E.~P. Simoncelli.
\newblock Image quality assessment: Unifying structure and texture similarity.
\newblock \emph{arXiv preprint arXiv:2004.07728}, 2020.

\bibitem[Fechner(1948)]{fechner1948elements}
Gustav~Theodor Fechner.
\newblock Elements of psychophysics, 1860.
\newblock 1948.

\bibitem[Feller(1991)]{feller1991introduction}
William Feller.
\newblock \emph{An introduction to probability theory and its applications, Volume 2}, volume~81.
\newblock John Wiley \& Sons, 1991.

\bibitem[Hepburn et~al.(2020)Hepburn, Laparra, Malo, McConville, and Santos{-}Rodr{\'{\i}}guez]{Hepburn2020perceptnet}
Alexander Hepburn, Valero Laparra, Jes{\'{u}}s Malo, Ryan McConville, and Ra{\'{u}}l Santos{-}Rodr{\'{\i}}guez.
\newblock Perceptnet: {A} human visual system inspired neural network for estimating perceptual distance.
\newblock In \emph{{IEEE} {ICIP} 2020}, pages 121--125. {IEEE}, 2020.

\bibitem[Jogan and Stocker(2014)]{jogan2014new}
Matja{\v{z}} Jogan and Alan~A Stocker.
\newblock A new two-alternative forced choice method for the unbiased characterization of perceptual bias and discriminability.
\newblock \emph{Journal of Vision}, 14\penalty0 (3):\penalty0 20--20, 2014.

\bibitem[Kingdom and Prins(2010)]{Kingdom10}
F.~A.~A. Kingdom and N.~Prins.
\newblock \emph{Psychophysics: A practical introduction}.
\newblock Elsevier Academic Press, 2010.

\bibitem[Laparra et~al.(2016)Laparra, Ball{\'e}, Berardino, and P]{laparra2016}
V.~Laparra, J~Ball{\'e}, A~Berardino, and Simoncelli~E P.
\newblock Perceptual image quality assessment using a normalized laplacian pyramid.
\newblock \emph{Electronic Imaging}, 2016\penalty0 (16):\penalty0 1--6, 2016.

\bibitem[Laparra et~al.(2011)Laparra, Camps-Valls, and Malo]{laparra2011iterative}
Valero Laparra, Gustavo Camps-Valls, and Jes{\'u}s Malo.
\newblock Iterative gaussianization: from ica to random rotations.
\newblock \emph{IEEE transactions on neural networks}, 22\penalty0 (4):\penalty0 537--549, 2011.

\bibitem[Laparra et~al.(2025)Laparra, Johnson, Camps-Valls, Santos-Rodríguez, and Malo]{laparra2025}
Valero Laparra, Juan~Emmanuel Johnson, Gustau Camps-Valls, Raúl Santos-Rodríguez, and Jesús Malo.
\newblock Estimating information theoretic measures via multidimensional gaussianization.
\newblock \emph{IEEE Transactions on Pattern Analysis and Machine Intelligence}, 47\penalty0 (2):\penalty0 1293--1308, 2025.

\bibitem[Larson and Chandler(2010)]{csiq-data}
E.~C. Larson and D.~M. Chandler.
\newblock Most apparent distortion: full-reference image quality assessment and the role of strategy.
\newblock \emph{JEI}, 19\penalty0 (1):\penalty0 011006, 2010.

\bibitem[Maloney and Yang(2003)]{maloney2003maximum}
Laurence~T Maloney and Joong~Nam Yang.
\newblock Maximum likelihood difference scaling.
\newblock \emph{Journal of Vision}, 3\penalty0 (8):\penalty0 5--5, 2003.

\bibitem[Perez-Ortiz and Mantiuk(2017)]{perez2017practical}
Maria Perez-Ortiz and Rafal~K Mantiuk.
\newblock A practical guide and software for analysing pairwise comparison experiments.
\newblock \emph{arXiv preprint arXiv:1712.03686}, 2017.

\bibitem[Ponomarenko et~al.(2009)]{tid2008-data}
N.~Ponomarenko et~al.
\newblock Tid2008-a database for evaluation of full-reference visual quality assessment metrics.
\newblock \emph{Advances of Modern Radioelectronics}, 10\penalty0 (4):\penalty0 30--45, 2009.

\bibitem[Ponomarenko et~al.(2013)]{tid2013-data}
N.~Ponomarenko et~al.
\newblock Color image database tid2013: Peculiarities and preliminary results.
\newblock In \emph{EUVIP}, pages 106--111. IEEE, 2013.

\bibitem[Sheikh et~al.(2005)Sheikh, Wang, Cormack, and Bovik]{live-data}
H.~R. Sheikh, Z.~Wang, L.~Cormack, and A.~C. Bovik.
\newblock Live image quality assessment database release 2.
\newblock \emph{http://live. ece. utexas. edu/research/quality}, 2005.

\bibitem[Silverstein and Farrell(2001)]{silverstein2001efficient}
D~Amnon Silverstein and Joyce~E Farrell.
\newblock Efficient method for paired comparison.
\newblock \emph{Journal of Electronic Imaging}, 10\penalty0 (2):\penalty0 394--398, 2001.

\bibitem[Thurstone(1994)]{thurstone1994law}
Louis~L Thurstone.
\newblock A law of comparative judgment.
\newblock \emph{Psychological review}, 101\penalty0 (2):\penalty0 266, 1994.

\bibitem[Toderici et~al.(2021)Toderici, Theis, Ballé, Johnston, Shi, Agustsson, Rapaka, Mentzer, Sinno, Norkin, Noury, and Timofte]{CLIC2021}
George Toderici, Lucas Theis, Johannes Ballé, Nick Johnston, Wenzhe Shi, Eirikur Agustsson, Krishna Rapaka, Fabian Mentzer, Zeina Sinno, Andrey Norkin, Erfan Noury, and Radu Timofte.
\newblock Workshop and challenge on learned image compression (clic2021), 2021.
\newblock URL \url{http://www.compression.cc}.

\bibitem[Tsukida et~al.(2011)Tsukida, Gupta, et~al.]{tsukida2011analyze}
Kristi Tsukida, Maya~R Gupta, et~al.
\newblock How to analyze paired comparison data.
\newblock \emph{Department of Electrical Engineering University of Washington, Tech. Rep. UWEETR-2011-0004}, 1, 2011.

\bibitem[Wang et~al.(2003)Wang, Simoncelli, and Bovik]{wang2003multiscale}
Z.~Wang, E.~P. Simoncelli, and A.~C. Bovik.
\newblock Multiscale structural similarity for image quality assessment.
\newblock In \emph{ACSSC}, volume~2, pages 1398--1402. Ieee, 2003.

\bibitem[Wang et~al.(2004)Wang, Bovik, Sheikh, and Simoncelli]{Wang04}
Zhou Wang, A.C. Bovik, H.R. Sheikh, and E.P. Simoncelli.
\newblock Image quality assessment: from error visibility to structural similarity.
\newblock \emph{IEEE Transactions on Image Processing}, 13\penalty0 (4):\penalty0 600--612, 2004.
\newblock \doi{10.1109/TIP.2003.819861}.

\bibitem[Zhang et~al.(2018)Zhang, Isola, Efros, Shechtman, and Wang]{zhang2018unreasonable}
R.~Zhang, P.~Isola, A.~Efros, E.~Shechtman, and O.~Wang.
\newblock The unreasonable effectiveness of deep features as a perceptual metric.
\newblock In \emph{Proceedings of the IEEE CVPR}, pages 586--595, 2018.

\end{thebibliography}

%%%%%%%%%%%%%%%%%%%%%%%%%%%%%%%%%%%%%%%%%%%%%%%%%%%%%%%%%%%%%%%%%%%%%%%%%%%%%%%
%%%%%%%%%%%%%%%%%%%%%%%%%%%%%%%%%%%%%%%%%%%%%%%%%%%%%%%%%%%%%%%%%%%%%%%%%%%%%%%
% APPENDIX
%%%%%%%%%%%%%%%%%%%%%%%%%%%%%%%%%%%%%%%%%%%%%%%%%%%%%%%%%%%%%%%%%%%%%%%%%%%%%%%
%%%%%%%%%%%%%%%%%%%%%%%%%%%%%%%%%%%%%%%%%%%%%%%%%%%%%%%%%%%%%%%%%%%%%%%%%%%%%%%
\newpage
\appendix
\onecolumn

\section{Perceptual Datasets}\label{ap:dataset}
Table~\ref{tab:datasets} shows the existing perceptual datasets, and what sort of experimental setup was used. Whilst most datasets use the 2AFC setup, some datasets (TID, CSIQ, CLIC) use an Elo ranking system to decide which images to show a particular observer. This results in a dataset where each triplet judgement is not independent, and results in triplets having a different number of judgements $M$.

BAPPS and CLIC are the only datasets that release the raw 2AFC ratings, but they differ in that BAPPS ensures the same number of judgements for each triplet, and each observer is shown random triplets. This is the setting that the proposed method was designed for, but we can still apply it to others. 

\begin{table}[htb]
\centering
\footnotesize
\begin{tabular}{llllllll}
\hline
Dataset & Method & \begin{tabular}[c]{@{}l@{}}Image\\ Sizes\end{tabular} & \begin{tabular}[c]{@{}l@{}}No. of\\ Images\end{tabular} & \begin{tabular}[c]{@{}l@{}}No. of\\ Distortions\end{tabular} & \begin{tabular}[c]{@{}l@{}}No. of\\ Triplets\end{tabular} & \begin{tabular}[c]{@{}l@{}}Total No. of\\ Judgements\end{tabular} & \begin{tabular}[c]{@{}l@{}}Type of\\ Judgement\\ Released\end{tabular} \\ \hline
TID 2008 & \begin{tabular}[c]{@{}l@{}}2AFC\\ sorting\end{tabular} & 512x384 & 25 & 17 & 2k & 256k & MOS \\
TID 2013 & \begin{tabular}[c]{@{}l@{}}2AFC\\ sorting\end{tabular} & 512x384 & 25 & 24 & 3k & 5k & MOS \\
CSIQ & 2AFC & 512x512 & 30 & 6 & 866 & 5k & DMOS \\
LIVE & \begin{tabular}[c]{@{}l@{}}5 level\\ scale\end{tabular} & 768x512 & 29 & 5 & 779 & 25k & DMOS \\
\begin{tabular}[c]{@{}l@{}}BAPPS\\ Train\end{tabular} & 2AFC & 64x64 & 151k & 425 & 151k & 302k & 2AFC \\
\begin{tabular}[c]{@{}l@{}}BAPPS\\ JND\end{tabular} & JND & 64x64 & 10k & 425 & 10k & 29k & True/False \\
\begin{tabular}[c]{@{}l@{}}BAPPS\\ Validation\end{tabular} & 2AFC & 64x64 & 36k & 425+ & 36k & 182k & 2AFC \\
CLIC 2021 & 2AFC & 768x768 & 315 &  & 119k & 120k & 2AFC \\ \hline
\end{tabular}
\vspace{0.3cm}
\caption{Detailed description of existing perceptual datasets: TID 2008\citep{tid2008-data}, TID 2013\citep{tid2013-data}, CSIQ~\citep{csiq-data}, LIVE~\citep{live-data}, BAPPS~\citep{zhang2018unreasonable} and CLIC~\citep{CLIC2021}. }
\label{tab:datasets}
\end{table}

\section{Marginal Uniformisation}\label{ap:uniform}
\begin{figure}[htb]
    \centering
    \includegraphics[width=\textwidth]{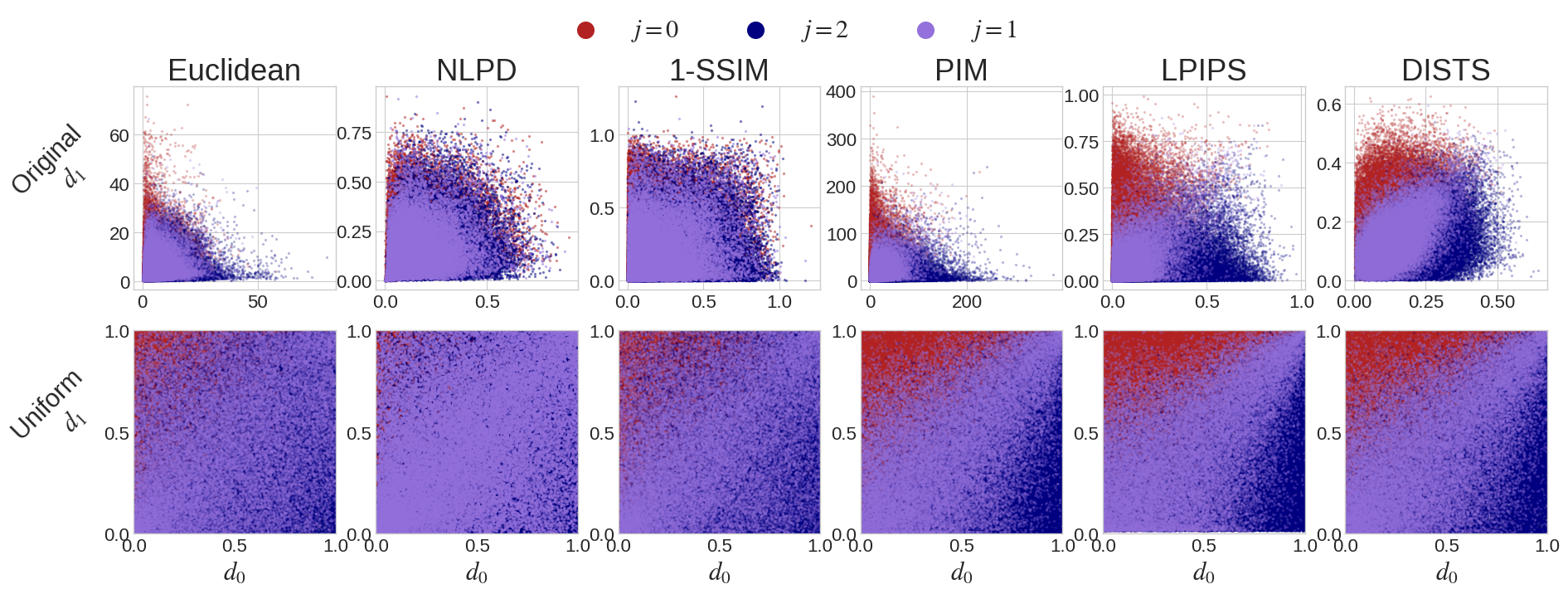}
    \vspace{0.3cm}
    \caption{Scatter plot of candidate distances in their original space (top row) and uniformised (bottom row). Shown are the training samples from the BAPPS dataset and the colour indicates the judgement assigned to triplet according to 2 observers.}
    \label{fig:enter-label}
\end{figure}

The most common approach to estimate a smooth function from a set of discrete events is to use adaptive kernels, with sizes depending on the density of values at a given point when estimating the density. Alternatively, we can search for transformations relocating the samples such that they cover the space fully and as evenly as possible. In this work, for each candidate distance, we transform the pairs of distances $\{ d_0, d_1\}$, such that the resulting set of points is marginally uniform in the range $[0, 1]$.
The marginal uniformisation transform is given by the histogram-equalisation solution
\begin{equation}
    u = U^i_{(k)}(x_i^{(k)}) = \int ^{x_i^{(k)}}_{-\infty} p_i({x'}_i^{(k)})d {x'}_i^{(k)}.
\end{equation}
where $x^{(k)}_i$ is a sample in the $k$th marginal and $p_i$ is approximated with histograms, for which computing the cumulative density function is straightforward~\citep {laparra2011iterative, laparra2025}. In doing so, we are transforming our data to a (fairly) uniform domain. This non-linear transformation facilitates the numerical density estimations from the discrete data but it is transparent for the posterior computations.
Fig.\ref{fig:enter-label} shows some examples of distances transformed to a uniform domain.

\section{Maximum likelihood density estimation}\label{ap:max_like}
We have that:
\begin{equation}\label{eq:bin_likelihood}
    Pr_{\mathcal{B}}(n=j;M,P) = \frac{M!}{j!(M-j)!}P^j (1-P)^{M-j},
\end{equation}
and we wish to minimise the likelihood of the obtained 2AFC answers having been generated from a binomial distribution with parameters $P(d_0,d_1)$ and $M$:
\begin{equation}\label{eq:lik}
    \begin{aligned}
    L(d_0,d_1;M,P) = \sum_{j=1}^{M}
    \big[ Pr(n=j|d_0,d_1) \cdot
    \log Pr_{\mathcal{B}}(n=j;M,P) \big].
    \end{aligned}
\end{equation}  
We use the latter expression in Eq.~\ref{eq:lik} for choosing the $P$ that maximises the likelihood of the empirical probabilities $\{Pr(n=j|d_0,d_1)\}$:
\begin{equation}\label{eq:maximised_likelihood2}
    \begin{split}
    \hat P(d_0,d_1) =&
        \arg \max_P \sum_{j=1}^{M} Pr(n=j|d_0,d_1)
        \left[
        \begin{aligned}
             \log&\left(\frac{M!}{j!(M-j)!}\right) + j\log P \\
            +& (M-j)\log(1-P)
        \end{aligned}
        \right]\\
        =&\frac{1}{M} \sum_{j=1}^{M} {j \cdot Pr(n=j|d_0,d_1)},
    \end{split}
\end{equation}
where the last equality is obtained by differentiating the argument of the right-hand side in the first equation w.r.t. $P$, equating to 0 and solving for $P$.

\section{Additional Evaluation} \label{ap:eval}
To maximise the agreement for the learned model, one can generate random samples $\hat{n}(t)$ from our learned distribution $\hat{n}(t) \sim \mathcal{B}(M, \hat{P}(d_0(t), d_1(t)))$, and see the agreement between the empirical and the simulated judgements, according to our learned model. This provides a reference for the case that the observed judgements follow the fitted binomial model exactly.

Similarly to the agreement of judgements, we can sample from our learned distribution with $M$ judgements $\hat{n}(t) \sim \mathcal{B}(M, \hat{P}(d_0(t), d_1(t)))$ and evaluate the negative log-likelihoods of these judgements, to get a minimum possible negative log-likelihood achieved by a distance.

Both of these measures are reported for samples generated from the learned models, evaluated against the training data, with $\text{AJ}(\hat{n}, \hat{P}, M)$ and $\text{NLL}(\hat{n}, \hat{P}, M)$.

\section{BAPPS Additional Results}\label{ap:bapps_results}
Here we present additional results on the training and test validation sets of BAPPS. We also separately report evaluation metrics per distortion used in the BAPPS test set for a more in-depth comparison of metrics.

Table~\ref{tab:ap_results_bapps} shows evaluation metrics on both the training and test set of BAPPS using the proposed density estimation method. The training set has been used to fit $\hat{P}(d_0, d_1)$. We see a consistent behaviour across sets, despite the different number of judgements $M$ for the train and test sets. Table~\ref{tab:ap_results_bapps_nn} reports the same measurements for the neural network method, displaying similar behaviour.

\begin{table}[htb]
\centering
\begin{tabular}{llllllll}
\hline
\multicolumn{1}{c}{Measure} & \multicolumn{1}{c}{} & \multicolumn{1}{c}{Euclidean} & \multicolumn{1}{c}{NLPD} & \multicolumn{1}{c}{SSIM} & \multicolumn{1}{c}{PIM} & \multicolumn{1}{c}{LPIPS} & \multicolumn{1}{c}{DISTS} \\ \hline
\multirow{2}{*}{$\text{AJ}(n, \hat{P}, M)$ (\%)$\uparrow$} & Train & $69.38$ & $69.16$ & $71.29$ & $80.52$ & $80.60$ & $79.65$ \\
 & Test & $75.89$ & $75.82$ & $76.17$ & $82.12$ & $82.43$ & $81.34$ \\ \hline
\multirow{2}{*}{$\text{AJ}(\hat{n}, \hat{P}, M)$ (\%)$\uparrow$ } & Train & $74.90$ & $74.26$ & $75.40$ & $81.85$ & $82.20$ & $81.23$ \\
 & Test & $83.02$ & $83.05$ & $83.25$ & $84.98$ & $85.37$ & $84.86$ \\ \hline
\multirow{2}{*}{NLL($n, \hat{P}, M$)$\downarrow$} & Train & $1.05$ & $1.07$ & $1.02$ & $0.76$ & $0.75$ & $0.78$ \\
 & Test & $1.86$ & $1.86$ & $1.84$ & $1.49$ & $1.46$ & $1.53$ \\ \hline
\multirow{2}{*}{NLL($\hat{n}, \hat{P}, M$)$\downarrow$} & Train & $0.93$ & $0.94$ & $0.91$ & $0.72$ & $0.71$ & $0.74$ \\
 & Test & $1.47$ & $1.47$ & $1.45$ & $1.31$ & $1.29$ & $1.33$ \\ \hline
\multirow{2}{*}{2AFC Score$\uparrow$} & Train & $66.80$ & $66.50$ & $68.88$ & $77.69$ & $77.05$ & $76.61$ \\
 & Test & $62.43$ & $63.78$ & $64.19$ & $70.31$ & $70.02$ & $68.81$ \\ \hline
\end{tabular}
\vspace{0.3cm}
\caption{Results on the BAPPS dataset~\citep{zhang2018unreasonable} using the density estimation method. Since this method is deterministic, we report only the value. In the training set, there are 2 judgements per triplet ($M=2$) and in the test set, 5 ($M=5$). Lower NLL is better. $\hat{n}$ denotes sampling judgements from the model optimised using the training data and evaluating against the true judgements.}
\label{tab:ap_results_bapps}
\end{table}

\begin{table}[htb]
\centering
\resizebox{\textwidth}{!}{%
\footnotesize
\begin{tabular}{llllllll}
\hline
\multicolumn{1}{c}{Measure} & \multicolumn{1}{c}{} & \multicolumn{1}{c}{Euclidean} & \multicolumn{1}{c}{NLPD} & \multicolumn{1}{c}{SSIM} & \multicolumn{1}{c}{PIM} & \multicolumn{1}{c}{LPIPS} & \multicolumn{1}{c}{DISTS} \\ \hline
\multirow{2}{*}{$\text{AJ}(n, \hat{P}, M)$ (\%)$\uparrow$} & Train & $69.28\pm0.12$ &$69.05\pm0.12$ &$71.20\pm0.16$ &$80.48\pm0.02$ &$80.59\pm0.01$ &$79.65\pm0.01$ \\
 & Test & $75.76\pm0.09$ &$75.79\pm0.08$ &$76.18\pm0.18$ &$82.07\pm0.13$ &$82.55\pm0.03$ &$81.40\pm0.03$ \\ \hline
\multirow{2}{*}{$\text{AJ}(\hat{n}, \hat{P}, M)$ (\%)$\uparrow$ } & Train & $74.96\pm0.30$ &$74.18\pm0.21$ &$75.54\pm0.55$ &$81.83\pm0.77$ &$82.51\pm0.17$ &$81.55\pm0.23$ \\
 & Test & $83.11\pm0.11$ &$82.95\pm0.06$ &$83.19\pm0.13$ &$85.05\pm0.36$ &$85.43\pm0.07$ &$85.00\pm0.11$ \\ \hline
\multirow{2}{*}{NLL($n, \hat{P}, M$)$\downarrow$} & Train & $1.05\pm0.00$ &$1.07\pm0.00$ &$1.02\pm0.00$ &$0.76\pm0.00$ &$0.75\pm0.00$ &$0.78\pm0.00$ \\
 & Test & $1.87\pm0.00$ &$1.86\pm0.00$ &$1.84\pm0.01$ &$1.50\pm0.01$ &$1.46\pm0.00$ &$1.53\pm0.00$ \\ \hline
\multirow{2}{*}{NLL($\hat{n}, \hat{P}, M$)$\downarrow$} & Train & $0.93\pm0.01$ &$0.94\pm0.01$ &$0.90\pm0.01$ &$0.72\pm0.03$ &$0.70\pm0.01$ &$0.73\pm0.01$ \\
 & Test & $1.46\pm0.01$ &$1.47\pm0.00$ &$1.45\pm0.01$ &$1.31\pm0.03$ &$1.28\pm0.00$ &$1.32\pm0.01$ \\ \hline
\multirow{2}{*}{2AFC Score$\uparrow$} & Train &  $66.69\pm0.06$ &$66.52\pm0.06$ &$68.90\pm0.10$ &$69.94\pm23.31$ &$77.09\pm0.05$ &$76.61\pm0.03$ \\
 & Test & $62.43\pm1.31$ &$63.78\pm1.20$ &$64.19\pm1.37$ &$63.28\pm21.13$ &$70.02\pm1.04$ &$68.81\pm0.49$ \\ \hline
\end{tabular}}
\vspace{0.3cm}
\caption{Results on the BAPPS dataset~\citep{zhang2018unreasonable} using a neural network. In the training set, there are 2 judgements per triplet ($M=2$) and in the test set, 5 ($M=5$). Lower NLL is better. $\hat{n}$ denotes sampling judgements from the model optimised using the training data and evaluating against the true judgements.}
\label{tab:ap_results_bapps_nn}
\end{table}

Table~\ref{tab:ap_distortion_bapps} shows a breakdown of the agreement of judgements (AJ) Eq.~\ref{eq:agreement}, negative log-likelihood (NLL) Eq.~\ref{eq:nll_likelihood} and 2AFC score Eq.~\ref{eq:2afc}, evaluated on the test set of BAPPS. We split the dataset into the category of distortion used, namely: Traditional (4720 triplets), CNN (4720 triplets), Color (9440 triplets), Deblur (1888 triplets), Frame interpolation (10856) and Super resolution (4720 triplets). Details on these distortions can be found in \cite{zhang2018unreasonable}.

\begin{table}[htb]
\centering
\footnotesize
\begin{tabular}{llllllll}
\hline
\multirow{2}{*}{Distance} & \multirow{2}{*}{Measure} & \multicolumn{6}{c}{Distortion} \\
 &  & Traditional & CNN & Color & Deblur & \begin{tabular}[c]{@{}l@{}}Frame\\ Interp.\end{tabular} & \begin{tabular}[c]{@{}l@{}}Super\\ Resolution\end{tabular} \\ \hline
\multicolumn{1}{c}{Euclidean} & $\text{AJ}(n, \hat{P}, M)$ (\%)$\uparrow$ & $65.54$ & $77.67$ & $78.33$ & $78.31$ & $75.93$ & $76.44$ \\
 & NLL($n, \hat{P}, M$)$\downarrow$ & $2.52$ & $1.72$ & $1.73$ & $1.71$ & $1.83$ & $1.82$ \\
 & 2AFC Score$\uparrow$ & $55.51$ & $80.72$ & $62.10$ & $57.81$ & $56.25$ & $66.18$ \\ \hline
\multirow{3}{*}{NLPD} & $\text{AJ}(n, \hat{P}, M)$ (\%)$\uparrow$ & $66.44$ & $76.78$ & $76.02$ & $78.32$ & $76.22$ & $77.15$ \\
 & NLL($n, \hat{P}, M$)$\downarrow$ & $2.48$ & $1.76$ & $1.84$ & $1.70$ & $1.81$ & $1.78$ \\
 & 2AFC Score$\uparrow$ & $56.61$ & $80.13$ & $59.24$ & $57.48$ & $55.72$ & $67.07$ \\ \hline
\multirow{3}{*}{SSIM} & $\text{AJ}(n, \hat{P}, M)$ (\%)$\uparrow$ & $67.86$ & $78.85$ & $76.05$ & $78.43$ & $77.38$ & $76.48$ \\
 & NLL($n, \hat{P}, M$)$\downarrow$ & $2.41$ & $1.64$ & $1.85$ & $1.70$ & $1.75$ & $1.82$ \\
 & 2AFC Score$\uparrow$ & $59.79$ & $80.83$ & $60.13$ & $58.49$ & $57.10$ & $65.01$ \\ \hline
\multirow{3}{*}{PIM} & $\text{AJ}(n, \hat{P}, M)$ (\%)$\uparrow$ & $81.44$ & $86.99$ & $80.25$ & $81.43$ & $82.02$ & $81.74$ \\
 & NLL($n, \hat{P}, M$)$\downarrow$ & $1.54$ & $1.15$ & $1.62$ & $1.55$ & $1.49$ & $1.51$ \\
 & 2AFC Score$\uparrow$ & $76.69$ & $83.75$ & $65.19$ & $62.15$ & $63.08$ & $71.53$ \\ \hline
\multirow{3}{*}{LPIPS} & $\text{AJ}(n, \hat{P}, M)$ (\%)$\uparrow$ & $80.58$ & $88.12$ & $80.90$ & $81.03$ & $81.62$ & $82.80$ \\
 & NLL($n, \hat{P}, M$)$\downarrow$ & $1.60$ & $1.06$ & $1.56$ & $1.56$ & $1.52$ & $1.44$ \\
 & 2AFC Score$\uparrow$ & $74.83$ & $83.64$ & $65.49$ & $61.39$ & $58.58$ & $69.84$ \\ \hline
\multirow{3}{*}{DISTS} & $\text{AJ}(n, \hat{P}, M)$ (\%)$\uparrow$ & $80.33$ & $85.92$ & $79.20$ & $80.17$ & $81.27$ & $81.77$ \\
 & NLL($n, \hat{P}, M$)$\downarrow$ & $1.61$ & $1.20$ & $1.66$ & $1.61$ & $1.54$ & $1.51$ \\
 & 2AFC Score$\uparrow$ & $75.47$ & $83.24$ & $63.96$ & $60.20$ & $62.55$ & $71.36$ \\ \hline
\end{tabular}
\vspace{0.3cm}
\caption{Evaluation metrics on the BAPPS validation set, split by distortion applied for the proposed density estimation method. Since the density estimation method is deterministic, we only report the value.}
\label{tab:ap_distortion_bapps}
\end{table}

\section{Interpretability - More Examples}\label{ap:int-examples}
\begin{figure}[htb]
        \centering
        \includegraphics[width=.8\textwidth]{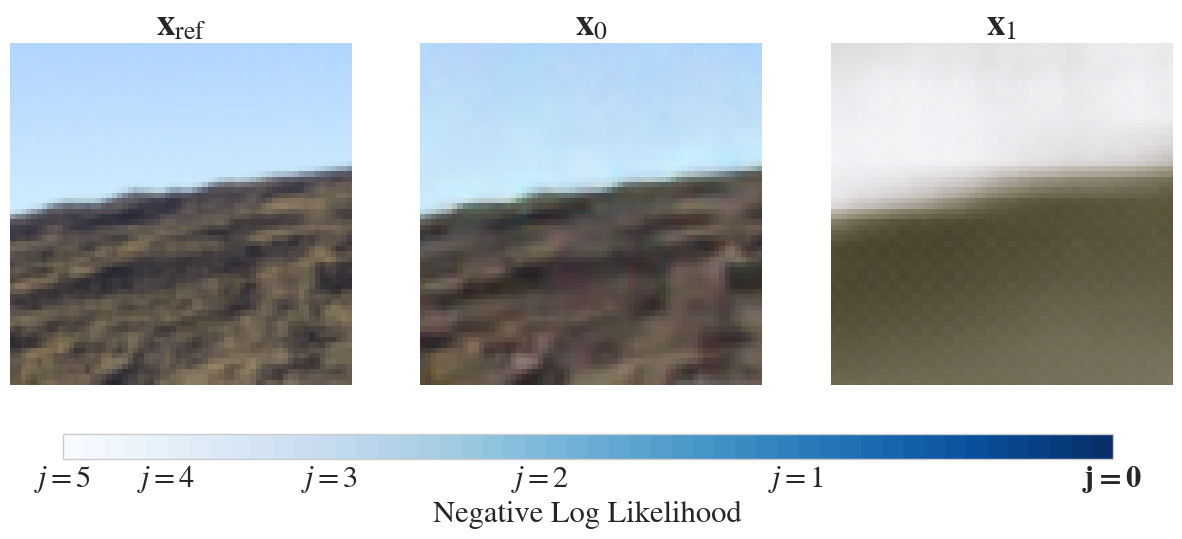}
        \vspace{0.3cm}
        \caption{Example of valuating the negative log-likelihood $j=[0, 5]$ according to DISTS for a triplet from the BAPPS test set where one image $\mathbf{x}_0$ is far from the reference $\mathbf{x}_{\text{ref}}$. White is more likely and blue is less likely.}
        \label{fig:int_eg2}
\end{figure}

\begin{figure}[htb]
        \centering
        \includegraphics[width=.8\textwidth]{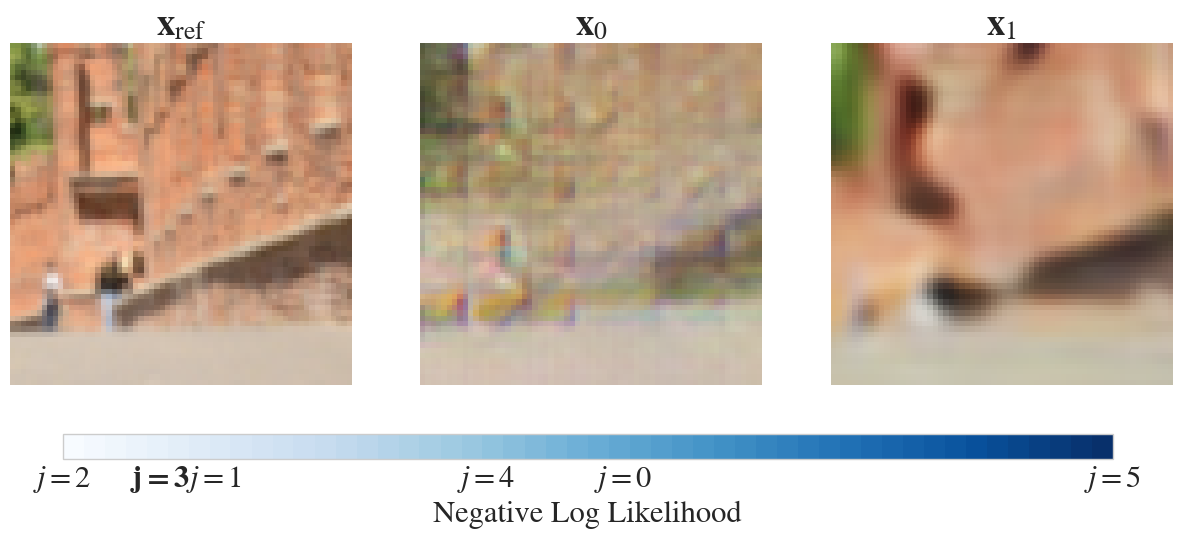}
        \vspace{0.3cm}
        \caption{Example of valuating the negative log-likelihood $j=[0, 5]$ according to DISTS for a triplet from the BAPPS test set where both images $\{\mathbf{x}_0, \mathbf{x}_1\}$ are far from the reference $\mathbf{x}_{\text{ref}}$. White is more likely and blue is less likely.}
        \label{fig:int_eg3}
\end{figure}

The negative log-likelihood in Eq~\ref{eq:nll_likelihood} depends on $\hat{P}(d_0, d_1)$, and in order to visualise this, here we present several examples of evaluating the NLL of different $j=[0, 5]$. We use triplets where one distorted image is extremely close to the reference and the decision is clear (Fig~\ref{fig:int_eg1}), one distorted image is extremely far from the original and the decision is clear (Fig~\ref{fig:int_eg2}) and finally a triplet where the decision is borderline as both distorted images are far from the original (Fig~\ref{fig:int_eg3}). For all experiments we use the density estimation method for estimation $\hat{P}(d_0, d_1)$

\section{Method Comparison}\label{ap:performance}
Table~\ref{tab:ap_compare} shows a comparison of the minimum training time, number of parameters, inference time of 1 sample (1 $(d_0, d_1)$ pair, and 128 samples. We report minimum training time as this gives a lower bound on how fast the method can train, accounting for other processes happening on the device. The density estimation method was ran on a Intel Xeon W-2245 CPU with 8 cores (16 with hyperthreading) and the neural network was trained on an NVIDIA RTX A5000 gpu.

\begin{table}[htb]
\centering
\resizebox{\textwidth}{!}{%
\begin{tabular}{@{}llllll@{}}
\toprule
Method & \begin{tabular}[c]{@{}l@{}}Minimum\\ Training time (s)\end{tabular} & \begin{tabular}[c]{@{}l@{}}No. of\\ Parameters\end{tabular} & \begin{tabular}[c]{@{}l@{}}Minimum Inference Time\\ 1 sample (s)\end{tabular} & \begin{tabular}[c]{@{}l@{}}Minimum Inference Time\\ 128 samples (s)\end{tabular} &  \\ \midrule
Density Estimation & 4.5 & 400 & $9.9 \times 10^{-8}$ & $3.7 \times 10^{-7}$ &  \\
Neural Network & 43.3 & 1281 & $1.0 \times 10^{-7}$ & $4.0 \times 10^{-7}$ &  \\ \bottomrule
\end{tabular}}
\vspace{0.3cm}
\caption{Comparison of minimum training time over 10 runs (giving a lower bound for training time) and number of learnable parameters for the proposed density estimation method and neural network counterpart. The density estimation method was ran on a Intel Xeon W-2245 CPU with 8 cores (16 with hyperthreading) and the neural network was trained on an NVIDIA RTX A5000 gpu.}
\label{tab:ap_compare}
\end{table}
\section{CLIC}\label{ap:clic}
We include additional information regarding the CLIC dataset used, including the distribution of the number of judgements $M_t$ per triplet. We also show additional visualisations of the triplets in the $(d_0, d_1)$ before and after the uniformistaion transformation, as well as evaluation metrics using both the training and test set.

\subsection{Distribution of Number of Judgements}
The CLIC 2021 subset we use to train (the oracle set) consists of 119,901 triplets with the number of judgements $M_t=\{1, 2\}$ and results of the judgements $j$, where the distribution can be seen in Fig.~\ref{fig:clic_train}. We also show the distributio of $j$ for each $M_t$. Most of the triplets have one judgement, with roughly uniform $j=\{0, 1\}$. For the triplets with 2 judgements, the majority are indecisive with $j=1$.

% \begin{figure}[htb]
%     \centering
%     \includegraphics[width=0.4\linewidth]{figs/CLIC_train_distributions.png}
%     \caption{Distribution of number of judgements $M_t$, and resulting judgements $j$ for the CLIC data used for training.}
%     \label{fig:clic_train_dist}
% \end{figure}

\begin{figure}[htb]
    \centering
    \includegraphics[width=0.99\linewidth]{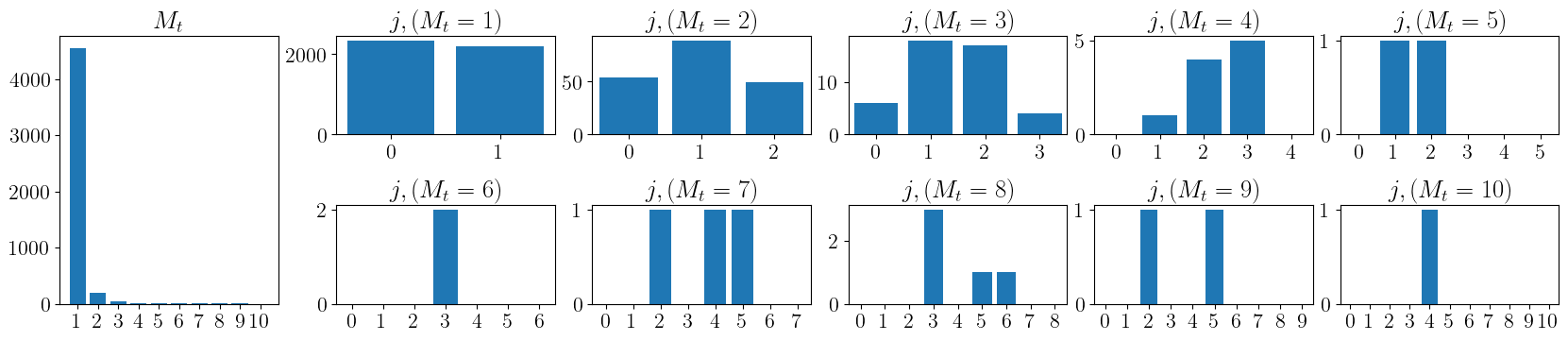}
    \vspace{0.3cm}
    \caption{Distribution of number of judgements $M_t$, and resulting judgements $j$ for the CLIC data used for evaluation.}
    \label{fig:clic_test_dist}
\end{figure}

The same distribution for the subset used for evaluation (the validation set)  with $M_t=[1, 10]$ can be seen in Fig~\ref{fig:clic_test_dist}. The vast majority of triplets also contain only 1 judgement, where the distribution of $j$ similar to that of the training set. The set also includes a small number of triplets with more judgements, varying in distribution of $j$.

\subsection{Additional Visualisations}
\begin{figure}[htb]
    \centering
    \includegraphics[width=\textwidth]{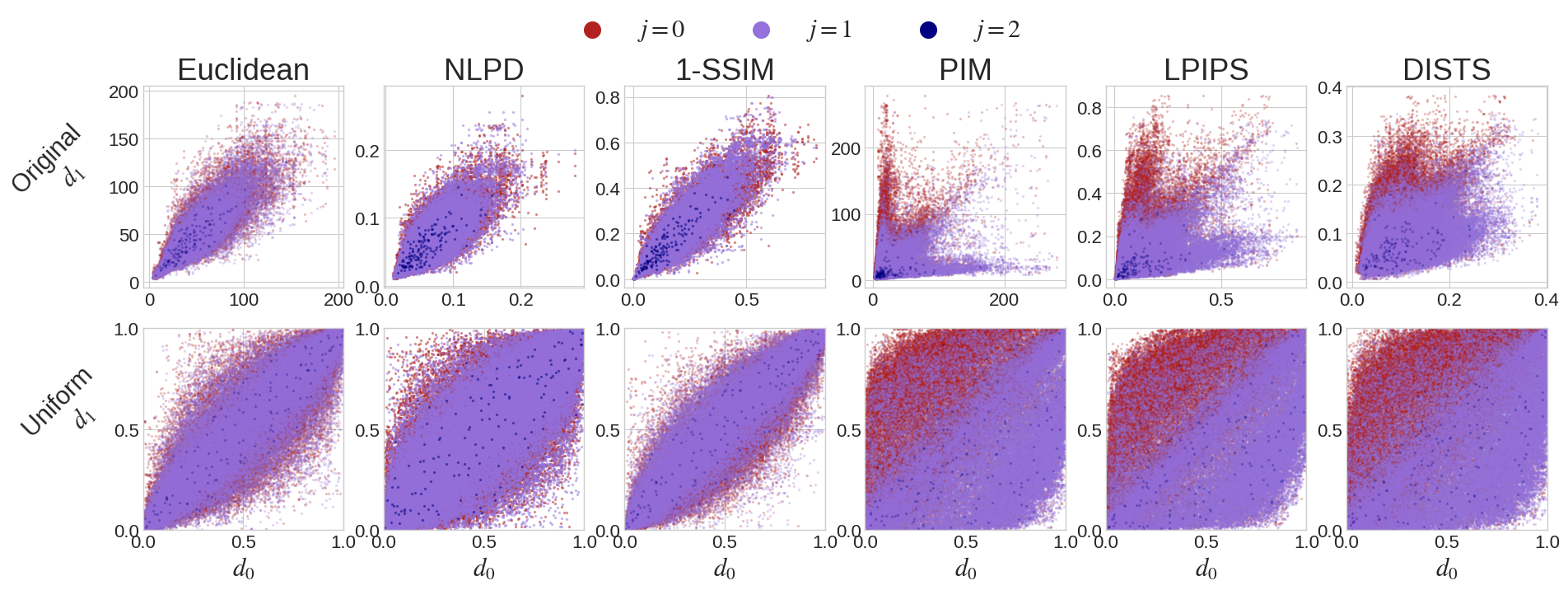}
    \vspace{0.3cm}
    \caption{Candidate distances in their original space (top row) and uniformised (bottom row). Shown are the training samples from the CLIC dataset and the colour indicates the judgement assigned to the triplet according to $\{1, 2\}$ observers. The points in this plot have a varying number of observers $M$.}
    \label{fig:clic_train}
\end{figure}

Fig~\ref{fig:clic_train} shows the distribution of triplets in the $(d_0, d_1)$ plane for the training set used to find $\hat{P}(d_0, d_1)$ from the CLIC dataset. Note that the triplets shown vary in number of judgements $M_t=\{1, 2\}$, and when training the triplets are treated as binary judgements ($M=1$) on $M_t$ identical triplets.

% \begin{figure}[htb]
%     \centering
%     \includegraphics[width=\textwidth]{figs/uniform_distances_clic_test.png}
%     \caption{Candidate distances in their original space (top row) and marginally uniformised (bottom row). Shown are the test samples from the CLIC dataset and the colour indicates the judgement assigned to the triplet according to 1 observers.}
%     \label{fig:clic_test}
% \end{figure}

We also show the surface of the binomial parameter $\hat{P}(d_0, d_1)$ in the $(d_0, d_1)$ plane estimated from the CLIC training set for both the proposed density estimation method and neural network.
\begin{figure}[htb]
    \centering
    \subfigure[Density Estimation]{%
        \includegraphics[width=0.9\textwidth]{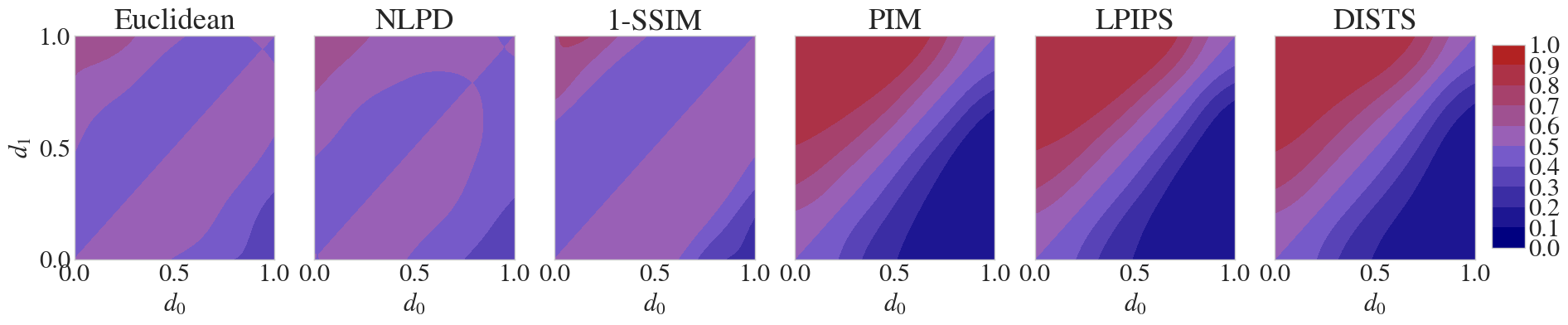}
    }
    \subfigure[Neural Network]{%
        \includegraphics[width=0.9\textwidth]{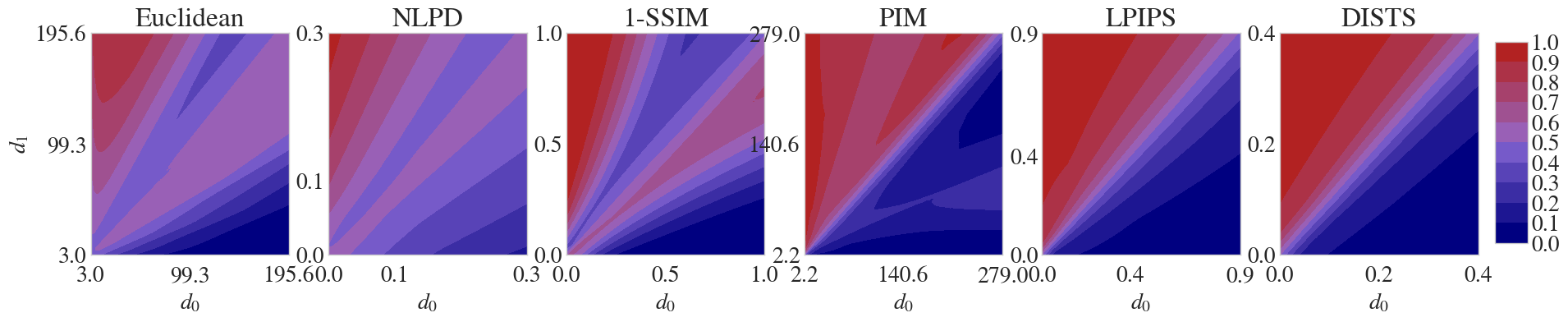}
    }
    \caption{Binomial parameter $P$ estimated from the training set of CLIC, for different candidate distances for (a) Density estimation and (b) Neural network.}
    \label{fig:decision_surface_clic}
\end{figure}

\end{document}